\documentclass[letterpaper,twocolumn,10pt]{article}
\usepackage{usenix2019_v3}

\usepackage{amsmath}
\usepackage{amssymb}
\usepackage[utf8]{inputenc}
\usepackage{amsmath}
\def \NAME {\textsc{ALTO}}

\newcommand{\sectionref}[1]{\hyperref[#1]{\S\ref*{#1}}}
\newcommand{\figureref}[1]{\hyperref[#1]{Fig. \ref*{#1}}}
\newcommand{\tabref}[1]{\hyperref[#1]{Table \ref*{#1}}}

\usepackage{enumitem}
\usepackage{lipsum}
\usepackage{listings}
\usepackage{xcolor}
\usepackage{tikz}
\usepackage{amsmath}
\usepackage{xparse}
\usepackage{caption}
\usepackage{setspace}
\usepackage{hyperref}
\usepackage[normalem]{ulem}

\usepackage{algorithmic}
\usepackage{algorithm}

\usepackage{graphicx}
\usepackage{caption}
\usepackage{subcaption}
\usepackage{svg}

\newcommand\commentone[1]{\textbf\textit{\textcolor{blue}{$\triangleright$~#1}}}

\lstdefinestyle{tt3}{
  basicstyle=\ttfamily\small,
  commentstyle=\color{gray},
  keywordstyle=\color{blue},
  breaklines=true,
  frame=single,
}

\definecolor{lightgray}{gray}{0.97}
\definecolor{codegray}{gray}{0.4}
\definecolor{codeblue}{rgb}{0.25,0.35,0.75}
\definecolor{codegreen}{rgb}{0,0.6,0}
\definecolor{codegray}{rgb}{0.5,0.5,0.5}
\definecolor{codepurple}{rgb}{0.58,0,0.82}
\definecolor{backcolour}{rgb}{0.95,0.95,0.92}
\definecolor{textblue}{rgb}{.2,.2,.7}
\definecolor{textred}{rgb}{0.54,0,0}
\definecolor{textgreen}{rgb}{0,0.43,0}
\definecolor{codered}{rgb}{201,72,12}

\definecolor{codegreen}{rgb}{0,0.6,0}
\definecolor{codegray}{rgb}{0.5,0.5,0.5}
\definecolor{codepurple}{rgb}{0.58,0,0.82}
\definecolor{backcolour}{rgb}{0.95,0.95,0.92}
\definecolor{textblue}{rgb}{.2,.2,.7}
\definecolor{textred}{rgb}{0.54,0,0}
\definecolor{textgreen}{rgb}{0,0.43,0}
\definecolor{codered}{rgb}{201,72,12}

\lstdefinestyle{ragdsl}{
language=Python,
basicstyle=\linespread{1}\ttfamily\footnotesize,
breaklines=true,
numbers=left,
frame=single,
numberstyle=\tiny, 
stepnumber=1,
numbersep=5pt, 
tabsize=4,
keywordstyle=\bfseries\color{codegreen},
commentstyle=\color{textred},   
stringstyle=\color{textgreen},
columns=fullflexible,
keepspaces=true,
xleftmargin=\parindent,
showstringspaces=false,
otherkeywords = {True, False},
keywordstyle=[2]\color{codepurple}\bfseries,
keywords=[2]{GNNAdvisor, GNNA},
keywordstyle=[3]\color{textblue}\bfseries,
keywords=[3]{__init__, forward},
keywordstyle=[4]\color{codegreen},
keywords=[4]{self},
}

\usepackage{soul,listings,xcolor}

\lstdefinestyle{tt3}{
  language=Python,
  basicstyle=\linespread{0.9}\ttfamily\footnotesize,
  breaklines=true,
  numbers=left,
  frame=single,
  numberstyle=\tiny,
  stepnumber=1,
  numbersep=5pt,
  tabsize=4,
  commentstyle=\color{textred},
  stringstyle=\color{textgreen},
  columns=fullflexible,
  keepspaces=true,
  xleftmargin=\parindent,
  showstringspaces=false,
  morecomment=[l]{\#},
  keywordstyle=[2]\color{codepurple}\bfseries,
  keywords=[2]{concat, cuBLAS, GroupedGEMM, FusedGEMM, BatchedGEMM},
  keywordstyle=[3]\color{textblue}\bfseries,
  keywords=[3]{X, W, A, B, S, Y, sched, base_out,
               dX, dY, dS, dX_base, dX_lora, dA, dB},
  keywordstyle=[4]\color{codegreen}\bfseries,
  keywords=[4]{T},
}

\lstdefinestyle{tt4}{
  language=Python,
  basicstyle=\linespread{0.9}\ttfamily\footnotesize,
  breaklines=true,
  numbers=left,
  frame=single,
  numberstyle=\tiny\color{gray},
  stepnumber=1,
  numbersep=5pt,
  tabsize=4,
  commentstyle=\color{textred},
  stringstyle=\color{textgreen},
  keywordstyle=[2]\color{codepurple}\bfseries,
  keywords=[2]{concat, cuBLAS, GroupedGEMM, FusedGEMM, BatchedGEMM, 
               Engine, 
               Task,
               create_task, EarlyExit, batched_execution},
  keywordstyle=[3]\color{textblue}\bfseries,
  keywords=[3]{X, W, A, B, S, Y, sched, base_out, dX, dY, dS, dX_base, 
               dX_lora, dA, dB, engine, best_adapters},
  keywordstyle=[4]\color{codegreen}\bfseries,
  keywords=[4]{alto, import, T},
  columns=fullflexible,
  keepspaces=true,
  xleftmargin=\parindent,
  showstringspaces=false,
  alsoletter={_},
  literate=
    *{0.10}{{\textcolor{textblue}{0.10}}}{4}
     {0.05}{{\textcolor{textblue}{0.05}}}{4}
     {1e-5}{{\textcolor{textblue}{1e-5}}}{4}
     {5e-6}{{\textcolor{textblue}{5e-6}}}{4}
}

\providecommand{\setcitestyle}[1]{}
\setcitestyle{nocompress}
\usepackage{booktabs}
\usepackage{tablefootnote}

\begin{document}
\title{{\NAME}: Adaptive LoRA Tuning and Orchestration for Heterogeneous LoRA Training Workloads}
\author{
{\rm Jingwei Zuo}$^{*\dagger}$,
{\rm Xinze Feng}$^{*\dagger}$,
{\rm Zien Liu}$^\dagger$,
{\rm Kaijian Wang}$^\dagger$,
{\rm Fanjiang Ye}$^\dagger$,
{\rm Ye Cao}$^\P$,\\
{\rm Zhuang Wang}$^\dagger$,
{\rm Yuke Wang}$^{\dagger}$\\[1ex]
$^\dagger$Rice University \quad $^\P$Independent Researcher
}


\date{}

\maketitle

{\let\thefootnote\relax\footnote{$^*$Jingwei Zuo and Xinze Feng contributed equally.}}
\begin{abstract}
{Low-Rank Adaptation (LoRA) is now the dominant method for parameter-efficient fine-tuning of large language models, but achieving a high-quality adapter often requires systematic hyperparameter tuning because LoRA performance is highly sensitive to configuration choices. In practice, this leads to many concurrent LoRA jobs, often spanning heterogeneous tasks in multi-tenant environments. Existing systems largely handle these jobs independently, which both wastes computation on weak candidates and leaves GPUs underutilized. We present \textbf{\NAME} (\textbf{\underline{A}}daptive \textbf{\underline{L}}oRA \textbf{\underline{T}}uning and \textbf{\underline{O}}rchestration), a co-designed training system that accelerates LoRA hyperparameter tuning while enabling efficient cluster sharing across heterogeneous tasks. The central insight behind {\NAME} is that when multiple tuning jobs run concurrently over a shared frozen backbone, they expose optimization opportunities that single-job designs cannot exploit. Building on this, {\NAME} monitors loss trajectories to terminate unpromising configurations early, uses fused grouped GEMM together with a new rank-local adapter parallelism to co-locate surviving adapters and reclaim freed GPU capacity, and combines intra-task and inter-task scheduling to improve multi-task placement by leveraging the predictable duration of LoRA jobs. Extensive evaluation shows that {\NAME} achieves up to $13.8\times$ speedup over state-of-the-art without sacrificing adapter quality.}
\end{abstract}

\section{Introduction}

Pre-trained Large Language Models (LLMs), such as Claude 3~\cite{anthropic2024claude3} and GPT-4~\cite{openai2023gpt4}, have shown strong performance across a broad range of tasks, including coding~\cite{chen2021evaluating, jain2024livecodebench}, multilingual understanding~\cite{shi2022language}, and conversational assistance~\cite{zheng2023judging, chiang2024chatbot}. 
The paradigm of pre-training followed by fine-tuning has enabled these models to further achieve state-of-the-art performance when adapted to personalized or domain-specific applications, such as legal analysis~\cite{guha2023legalbench}, personalized education~\cite{chen2024gptutor}, and customer support~\cite{wulf2024customer}.

However, fine-tuning large models with all model parameters updated poses a significant challenge due to the high computational cost of training and serving numerous fine-tuned variants.
To address this, parameter-efficient fine-tuning (PEFT) techniques have emerged as scalable alternatives to traditional full fine-tuning, among which Low-Rank Adaptation (LoRA)~\cite{hu2022lora} is particularly popular due to its simplicity and effectiveness.
LoRA significantly reduces the number of trainable parameters by introducing low-rank decomposition matrices into model layers, allowing for specialization while keeping the pre-trained model weights frozen. 

\begin{figure}[t]
    \centering
    \includegraphics[width=\linewidth]{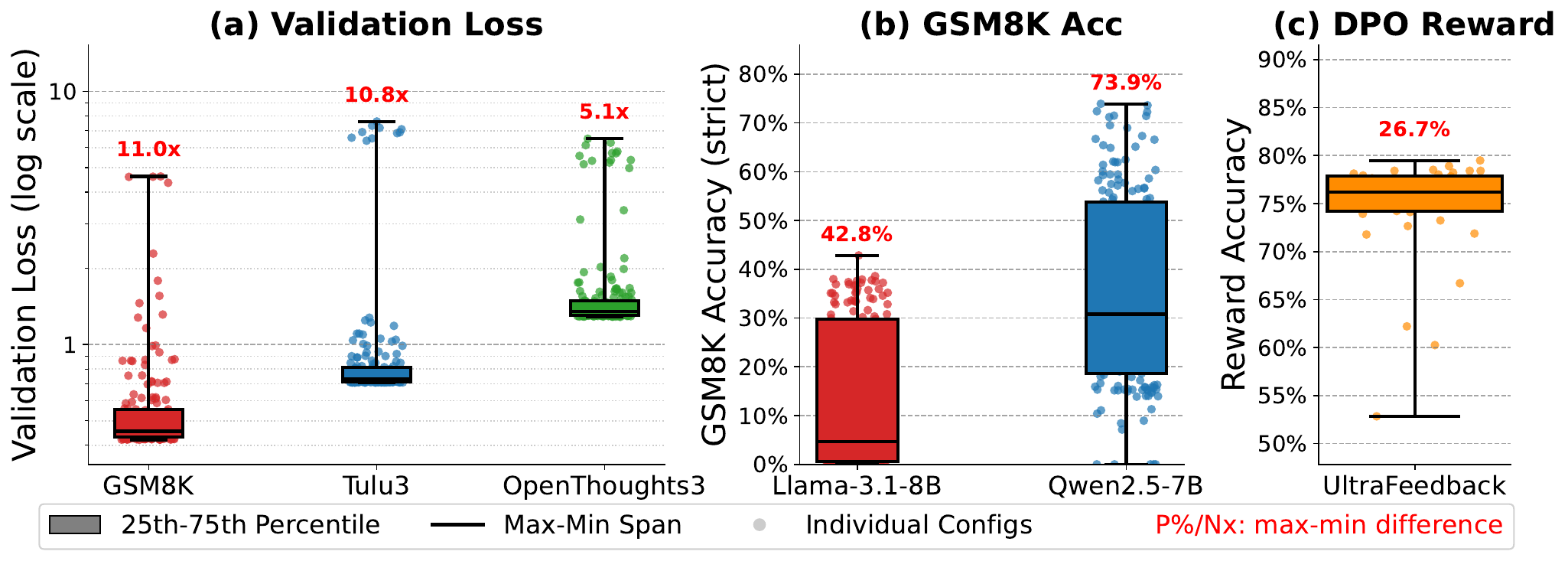}
    \caption{Different hyperparameters yield significantly varying downstream task performance in LoRA training. (a) Best validation loss distributions across 165 hyperparameter configs (varying batch size, learning rate, and LoRA rank) for Llama-3.1-8B. (b) GSM8K test accuracy of the best checkpoint per config shows a maximum of 73.9\% difference.  (c) DPO Reward accuracy of 60 different configurations on UltraFeedback dataset shows a maximum of 26.7\% difference. } 
    \label{fig:val_loss_stats}
\end{figure}

\begin{figure*}[t]
  \centering
  \includegraphics[width=\textwidth]{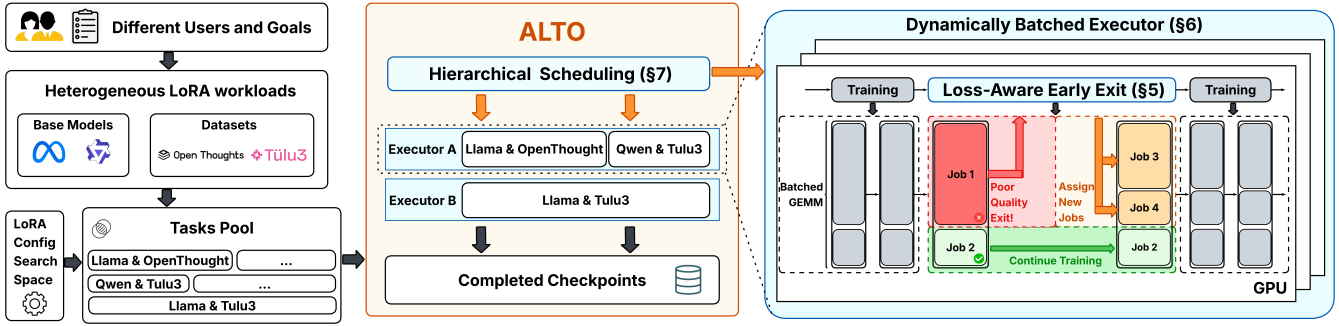}
  \caption{System Overview of {\NAME}.}
  \label{fig:sys-overview}
\end{figure*}

{Due to these substantial resource savings}, LoRA fine-tuning has been widely adopted in both large-scale cloud platforms~\cite{fireworks2024, together, tinker2025} and resource-constrained local environments~\cite{unsloth, llamafactory} for LLM post-training methods, including supervised fine-tuning (SFT)~\cite{ouyang2022training} and reinforcement learning (RL)~\cite{sheng2025hybridflow}.
Specifically, a LoRA fine-tuning \emph{\underline{task}} takes as input a base model and a training dataset, and aims to produce a high-quality LoRA adapter, as measured by task-specific metrics such as accuracy in SFT or reward in RL. 
Similarly to model pre-training, LoRA fine-tuning also requires specifying hyperparameters, such as learning rate, batch size, and adapter rank.
We denote a LoRA fine-tuning \emph{\underline{job}} as the training of a LoRA adapter under a \textit{specific} hyperparameter configuration. 
A LoRA task may consist of one or more fine-tuning jobs.

Although a LoRA fine-tuning job offers substantial efficiency gains over full parameter fine-tuning, how to efficiently train a LoRA task to obtain a decent LoRA adapter is challenging.
Our large-scale empirical study reveals the necessity of hyperparameter search in a LoRA task, and there is no single rule of thumb for hyperparameter tuning (\S\ref{sec:motivation}).
We conducted extensive experiments on hyperparameter sensitivity in LoRA fine-tuning across various model architectures, model sizes, and downstream tasks in both SFT and RL. 
As shown in Figure~\ref{fig:val_loss_stats}, the adapter quality gap between the best and worst configurations can be up to 73.9\%.
Unfortunately, existing LoRA fine-tuning frameworks, such as PEFT~\cite{peft}, LLamaFactory~\cite{llamafactory} or Unsloth~\cite{unsloth}, are unaware of this hyperparameter sensitivity and instead treat the LoRA task as a passive and sequential execution of provided jobs.
Consequently, practitioners have to manually launch a sequence of LoRA fine-tuning jobs over a hyperparameter search space to obtain a high-quality LoRA adapter for a given task, making the task-complete time increase linearly with the search space.
Additionally, because interests and goals often differ between tasks, this pattern of multiple jobs naturally extends to multiple tasks submitted by a single user or many users in shared training environments such as cloud platforms~\cite{lin2025lobra,ye2024mlora,googlevertexai2024,awssagemaker2024,azureopenai2024}.

To support these multi-job and multi-task workloads, existing systems~\cite{peft, llamafactory, unsloth} typically execute LoRA jobs separately, often by training one adapter at a time and switching the active adapter between jobs. Because individual LoRA jobs are lightweight, this approach results in low hardware utilization and, when scaled across multiple GPUs, disproportionately high distributed overhead.
A common optimization strategy is multi-LoRA fine-tuning~\cite{ye2024mlora,yan2025plora,lorafusion,tlora}, which runs multiple LoRA jobs jointly on the same GPUs. However, they still operate at job granularity and do not account for heterogeneity across tasks and job progress, leaving substantial GPU resources underutilized in multi-task workloads.

To explore optimization opportunities to address these inefficiencies, we identify several key observations. \underline{First}, LoRA hyperparameter quality is highly uneven across candidates over the course of training and does not reliably transfer between tasks. Within each LoRA task, many underperforming jobs can be identified early using objective-aligned training metrics, such as the supervised loss in SFT, because these signals directly measure progress toward the task objective~\cite{prechelt2002early}. This motivates a dynamic, performance-aware system design that improves training efficiency while ensuring strong downstream LoRA performance. \underline{Second}, we observe a fundamental conflict between the statistical behavior and system efficiency of LoRA fine-tuning. Specifically, our analysis (\S\ref{sec:motiv_early_exit}) suggests that small batch sizes are statistically preferred for strong LoRA convergence, whereas sufficiently large batch sizes are preferred in practice for high hardware utilization. Moreover, in multi-GPU settings, common distributed strategies either do not support batch sizes smaller than the world size~\cite{zhao2023pytorch,rajbhandari2020zero} or suffer from pipeline bubbles~\cite{pp1, pp2, pp3} and frequent communication overhead~\cite{shoeybi2019megatron, narayanan2021efficient}. Addressing this mismatch requires a coordinated approach that jointly optimizes convergence quality and hardware efficiency. {\underline{Third}, although realistic LoRA workloads are heterogeneous along many dimensions, such as training dataset size, their execution durations can often be estimated reliably before training begins. This predictability creates optimization opportunities for workload scheduling to reduce idle system resources.}

\begin{figure}[t]
  \small
\begin{lstlisting}[caption={Example of Using {\NAME} API.}, style=tt4, label={lst:alto}]
import alto
# 1. Initialize engine
engine = alto.Engine(strategy="adapter_parallel", total_gpus=8)
# 2. Define and batch heterogeneous tasks
tasks = [
  alto.Task(
      model="meta-llama/Llama-3.1-70B", num_gpus=4,
      dataset="math/gsm8k",
      search_space={"lr": [1e-5], "batch_size": [1, 2]},
  ),
  # ... Different model, dataset and search space
]
# 3. Set early-exit strategy, schedule and execute
early_exit_strategy = alto.EarlyExit(warmup_ratio=0.10)
schedule = engine.schedule(tasks, method="MILP")
best_adapters = engine.batched_execution(tasks, schedule, early_exit_strategy)
\end{lstlisting}
\vspace{-20pt}
\end{figure}

Based on these insights, we build \textbf{\NAME} (\textbf{\underline{A}}daptive \textbf{\underline{L}}oRA \textbf{\underline{T}}uning and \textbf{\underline{O}}rchestration, as illustrated in Figure~\ref{fig:sys-overview}), a unified LoRA training system that jointly adapts hyperparameter selections, hardware execution strategy, and scheduling to efficiently serve heterogeneous post-training tasks for LLMs. {\NAME} introduces three coordinated techniques that address the above challenges across statistical, system, and scheduling dimensions. \underline{First}, to avoid wasting compute on unpromising candidates, {\NAME} employs a \textit{loss-aware early-exit} mechanism that leverages the signals of training dynamics to terminate poor configurations after only a fraction of total training steps, freeing resources for the most promising trials. \underline{Second}, to resolve the tension between the small batch sizes favored by LoRA algorithm and the large batch sizes needed for hardware efficiency, {\NAME} introduces \textit{adapter batching} with \textit{adapter parallelism}: multiple small-batch LoRA jobs are co-located on a shared frozen backbone via fused grouped GEMM kernels, recovering GPU utilization without inflating per-adapter batch size, and scaling across devices by assigning each rank a distinct adapter rather than a distinct micro-batch. \underline{Third}, to orchestrate the resulting dynamic mix of jobs and tasks, {\NAME} adopts a \textit{hierarchical scheduling} strategy: an offline inter-task scheduler that exploits the predictable duration of LoRA fine-tuning tasks to plan placement and minimize end-to-end makespan, paired with an online intra-task scheduler that seamlessly supports dynamic LoRA job onloading after early exits. We exhibit an example (Listing~\ref{lst:alto}) that illustrates the typical usage of {\NAME}. End users can seamlessly integrate our frameworks with their existing LoRA-finetuning workflow with minimum additional efforts while enjoying from maximum efficiency gains.

To sum up, we make the following contributions:

\begin{itemize}

    \item We establish the necessity of hyperparameter tuning for LoRA across both SFT and RL workloads~(\S\ref{sec:motivation}), and introduce an adaptive early-exit mechanism that leverages training dynamics to reduce tuning cost~(\S\ref{sec:early-exit}).




    \item We build {\NAME}, a performance-aware LoRA training engine that combines the above mechanism with dynamic adapter batching as well as adapter parallelism to reduce the runtime cost of fine-tuning tasks~(\S\ref{sec:batched-engine}).

    \item To further scale {\NAME} under heterogeneous multi-task workloads, we propose a hierarchical scheduling via dynamic task placement and online job admission upon early exits to reduce idle GPU resources and end-to-end makespan~(\S\ref{sec:scheduling}).


    \item Extensive evaluation shows that  {\NAME} achieves up to 13.8$\times$ improvement in terms of training makespan compared to the existing state-of-the-art ~(\S\ref{sec:evaluation}).
    
    
    
\end{itemize}
\section{Background}
\label{sec:background}

This section provides essential background on LoRA fine-tuning of LLMs and distributed training parallelism.

\subsection{LoRA Fine-Tuning of Large Language Models}
Modern large language models (LLMs) are developed in two broad phases: \emph{pre-training}, which learns general-purpose representations from large-scale corpora, and \emph{fine-tuning}, which aligns the model to desired behaviors and downstream tasks.
Full-model fine-tuning updates all parameters and incurs substantial memory overhead for storing gradients and optimizer states, which becomes prohibitive as model scale grows. Parameter-Efficient Fine-Tuning (PEFT) methods~\cite{houlsby2019adapters, lester2021prompt, li2021prefix, liu2022ptuning} address this by freezing the pre-trained weights and introducing a small number of trainable parameters.
Among PEFT methods, Low-Rank Adaptation (LoRA)~\cite{hu2022lora} has emerged as the dominant approach. For a frozen weight matrix $W \in \mathbb{R}^{k \times n}$ in a target linear layer, LoRA injects a low-rank residual branch so that the output becomes $Y = XW + XAB$, where $A \in \mathbb{R}^{k \times r}$ and $B \in \mathbb{R}^{r \times n}$ are the only trainable matrices, with rank $r \ll \min(k, n)$. In practice, LoRA adds fewer than 1\% additional parameters relative to the base model~\cite{hu2022lora, dettmers2023qlora}.
LoRA's simplicity and low resource footprint have driven widespread adoption across both cloud fine-tuning platforms~\cite{together, anyscale, lambda} and local training toolkits~\cite{llamafactory, peft}, making it the de facto method for adapting LLMs to custom tasks.

Recently, there has been growing demand for LoRA training, which is heterogeneous both in backbone model and datasets. Enterprise spending on LLM fine-tuning grew 2.5$\times$ from 2023 to 2024, reaching an average of \$18 million per organization~\cite{a16z2024generativeai}, yet most organizations lack the ML expertise and GPU infrastructure to conduct LoRA training independently.
This expertise gap is compounded by the heterogeneity of fine-tuning requirements: users need to adapt diverse base models (e.g., LLaMA, Mistral, Qwen) to domain-specific datasets spanning healthcare, finance, law, and code generation, each with different rank configurations and resource constraints. Major cloud providers have already validated the Training-as-a-Service model: Azure OpenAI~\cite{azureopenai2024}, AWS SageMaker~\cite{awssagemaker2024}, and Google Vertex AI~\cite{googlevertexai2024} all offer managed LoRA-based fine-tuning APIs. Meanwhile, frameworks such as LoRAX~\cite{predibase2024lorax} demonstrate that thousands of LoRA adapters can be served from a single GPU at minimal overhead. These trends collectively underscore the need for efficient, scalable systems that provide LoRA training as an elastic, multi-tenant service. LoRA fine-tuning can be applied to not only \emph{supervised fine-tuning} (SFT), but also \emph{reinforcement learning from human feedback} (RLHF), according to recent research~\cite{schulman2025lora, wang2025tina}. Among mainstream RLHF methods~\cite{schulman2017proximal, rafailov2023direct, guo2025deepseek}, we pick direct preference optimization (DPO)~\cite{rafailov2023direct} as our demonstration of the application of LoRA training on RLHF processes.
\subsection{Distributed Parallelism Strategies}
Scaling training to multiple GPUs relies on several parallelism strategies. Data parallelism (DP)~\cite{abadi2016tensorflow, li2020pytorch} replicates the model across GPUs and partitions the training data, synchronizing gradients after each step. Fully Sharded Data Parallelism (FSDP)~\cite{zhao2023pytorch, rajbhandari2020zero} reduces memory by sharding parameters, gradients, and optimizer states across ranks. Tensor parallelism (TP)~\cite{shoeybi2019megatron, narayanan2021efficient} partitions individual linear layers across GPUs, requiring intra-layer all-reduce communication at each layer boundary. Pipeline parallelism (PP)~\cite{pp1, pp2, pp3} divides the model into sequential stages assigned to different GPUs, reducing per-device memory but introducing bubbles.

These strategies were designed for large-scale pre-training and become mismatched with LoRA fine-tuning, where trainable adapters constitute fewer than 1\% of total parameters and small batch sizes are statistically preferred (\S\ref{sec:motivation}). TP adds per-layer all-reduce communication whose latency dwarfs the microsecond-scale low-rank GEMMs. PP suffers from pipeline bubbles that worsen as micro-batch count shrinks, leaving most stages idle when per-adapter batch sizes approach~1. Multi-LoRA \emph{serving} frameworks such as S-LoRA~\cite{sheng2024slora} and Punica~\cite{chen2024punica} batch inference across adapters on a shared backbone, but their decode-phase kernels (BGMV/SGMV) operate at vector granularity and lack backward-pass support, making them inapplicable to training.
Among these strategies, FSDP has become the de facto standard for distributed LoRA training. LoRA fine-tuning with small batch sizes (\S\ref{sec:motivation}) exposes fundamental inefficiencies: FSDP shards the entire model across $P$ ranks but replicates the data pipeline, so when the global batch size is smaller than $P$, some ranks receive zero samples and become idle for LoRA computation while still paying the full all-gather and reduce-scatter cost for base-model weights. Moreover, since LoRA fine-tuning is dominated by the cost of loading base model weights from HBM to SRAM~\cite{lorafusion}, FSDP's redundant replication of LoRA adapter matrices across ranks further exacerbates the inefficiency. None of these strategies exploit LoRA's distinguishing characteristic: the trainable parameters are negligibly small compared to the frozen backbone, creating an opportunity for a parallelism design that shards the backbone for memory while keeping adapters local for compute.
\section{Motivation}
\label{sec:motivation}
In this section, we identify three key observations in providing efficient LoRA training services.

\label{sec:motiv_early_exit}



\textbf{Observation 1: LoRA HP tuning is necessary but incurs substantial redundancy.} Downstream performance of LoRA training for LLMs critically depends on selecting appropriate hyperparameters---including learning rate, batch size, LoRA rank, etc, so that they must be tuned for every single LoRA finetuning task.
We pick two commonly used LLMs for LoRA finetuning: Llama-3.1-8B~\cite{grattafiori2024llama} and Qwen2.5-7B~\cite{qwen2025qwen25technicalreport}, and three datasets on different fields: GSM8K~\cite{cobbe2021training} for math, Tulu-3~\cite{lambert2024tulu} for instruction following, and OpenThoughts3~\cite{guha2506openthoughts} for general reasoning.
As shown in Figure~\ref{fig:val_loss_stats} (a-c), the validation loss\footnote{Tulu-3 and OpenThoughts3 don't have ground truth answers, so we fall back to the validation loss for the evaluation of the downstream task performance.} varies dramatically across hyperparameter configurations, with the gap between the best and worst configurations exceeding an order of magnitude. Meanwhile, for the GSM8K dataset, we use vLLM \cite{kwon2023efficient} to evaluate the best checkpoint of each hyperparameter configuration picked by validation loss, and compare between different hyperparameter configurations. Figure~\ref{fig:val_loss_stats} (b) shows that the best hyperparameter configurations yield 42.8\% and 73.9\% accuracy on the evaluation dataset respectively, while the worst often have 0\% accuracy, and the median accuracy lies at 4\% and 31\%.

Furthermore, we evaluate the impact of hyperparameters on preference alignment using Direct Preference Optimization (DPO) with the Qwen2.5-32B~\cite{qwen2025qwen25technicalreport} model on the UltraFeedback~\cite{cui2023ultrafeedback} dataset. As illustrated in Figure~\ref{fig:val_loss_stats} (c), the reward accuracy exhibits a significant spread of 26.7\% between the optimal and suboptimal configurations. Even for larger models and alignment objectives, poor hyperparameter choices can cause reward accuracy to drop from nearly 80\% to approximately 53\%.
With all these experiments, we proved that \emph{the hyperparameter tuning is a must for LoRA training.} Thus, we are motivated to include hyperparameter tuning process in our system to yield the best adapter for each user's training task.

However, a na\"ive hyperparameter grid search incurs substantial redundancy, as it launches each configuration as an independent training job and runs it to completion before selecting the best. The redundancies are three-fold: 
(i)~some configurations never converge due to misconfigured settings (e.g., excessively large learning rates);
(ii)~even for configurations that do converge, the optimal checkpoint may occur well before the end of training, with later steps only degrading performance through overfitting;
and (iii)~since only a single best adapter is ultimately needed, all resources spent on configurations that are clearly underperforming are wasted.
In current practice, identifying these redundancies requires manual inspection of training curves, which demands continuous human attention, scales poorly with the number of concurrent configurations, and inevitably introduces reaction delays between the emergence of an ill pattern and its termination. This overhead becomes prohibitive in a service setting where dozens to hundreds of configurations may be training simultaneously across multiple tasks. This motivates an early-exit mechanism to reduce training redundancies. However, because training dynamics are different across dataset and models, this is challenging and requires sophisticated designs. 




\begin{figure}[t]
  \centering
  \includegraphics[width=\linewidth]{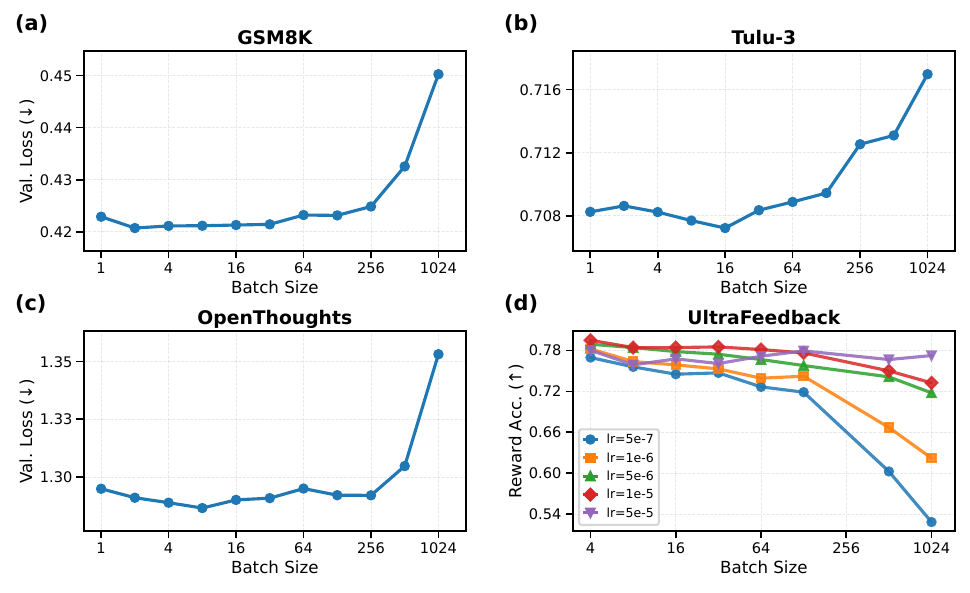}
  \vspace{-2.5em}
  \caption{LoRA fine-tuning prefer small batch size training. SFT validation loss for Llama-3.1-8B on (a) GSM8K, (b) Tulu-3, and (c) OpenThoughts3; 
   DPO reward accuracy for Qwen2.5-32B on (d) UltraFeedback across five learning rates. Performance peaks at small batch sizes ($\leq$16) and degrades beyond 32 for these settings.}
  \label{fig:batch-size-effect}
\end{figure}


\textbf{Observation 2: A fundamental conflict between the statistical behavior and system efficiency of LoRA fine-tuning.} 
Typically, machine learning practitioners increase the training batch size to improve GPU utilization during hyperparameter search.
However, we find that small batch sizes are often \emph{preferable} in terms of adapter quality for LoRA training.
Figure~\ref{fig:batch-size-effect}~(a-c) shows the validation loss across a range of batch sizes for Llama-3.1-8B on three SFT datasets: smaller batch sizes consistently achieve superior results, while excessively large batch sizes significantly degrade final performance.
This finding extends beyond SFT: Figure~\ref{fig:batch-size-effect}~(d) shows that for direct preference optimization (DPO) training~\cite{rafailov2023direct} of Qwen2.5-32B~\cite{qwen2025qwen25technicalreport} on UltraFeedback~\cite{cui2023ultrafeedback}, the adapter achieves the highest reward accuracy at small batch sizes across all learning rates.
Similar conclusions are also reported in prior studies ~\cite{schulman2025lora, lee2026bewarebatchsizehyperparameter}.

\begin{figure}[t]
    \centering
    \includegraphics[width=\linewidth]{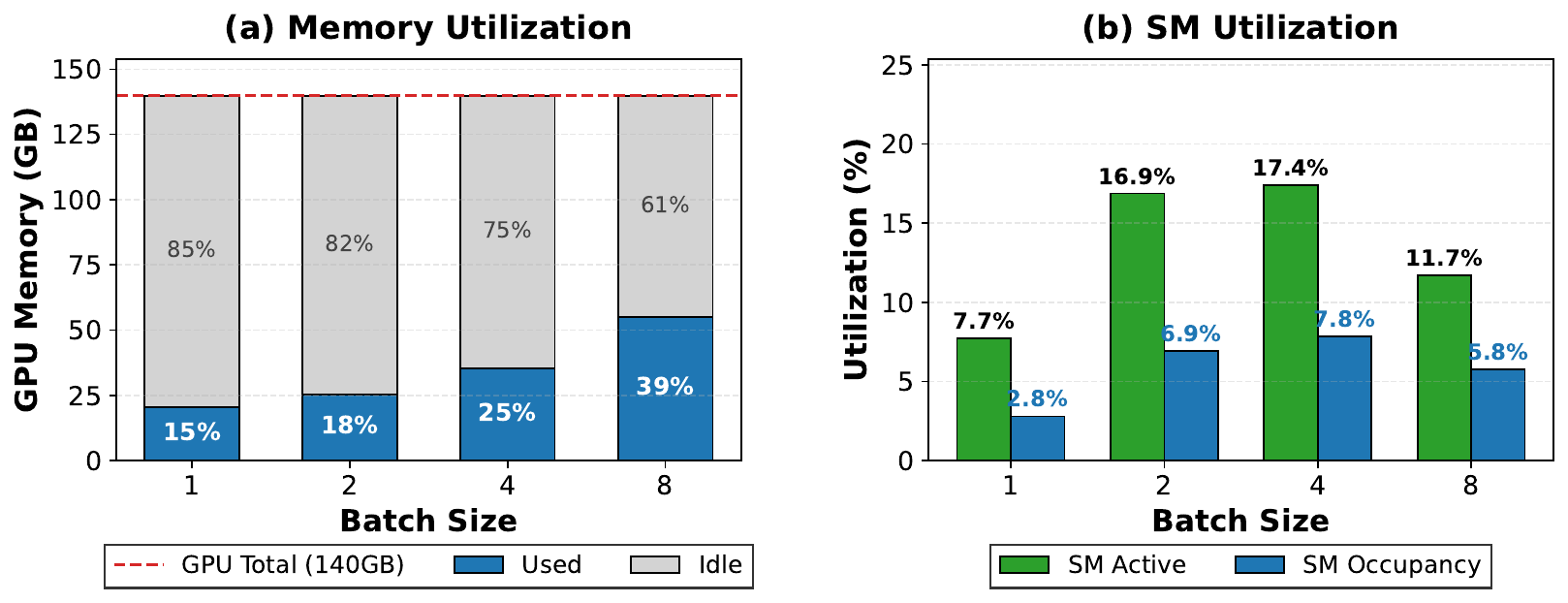}
    \caption{GPU memory and average SM utilization when training a single LoRA adapter with small batch sizes. 
    A substantial portion of GPU resources remains idle, motivating the need for batched multi-adapter training.}
    \label{fig:gpu_util}
\end{figure}

This algorithmic preference for small batches creates a tension with hardware efficiency.
On a single device, training a single LoRA adapter with a small batch severely underutilizes both GPU memory capacity and streaming multiprocessor (SM) occupancy (Figure~\ref{fig:gpu_util}).
The problem is further exacerbated in the multi-device setting.
When the base model exceeds single-GPU memory, distributed training becomes necessary.
Fully-Sharded Data Parallelism (FSDP)~\cite{zhao2023pytorch}, the standard approach, shards model parameters across GPUs and replicates the data pipeline.
However, FSDP is a particularly poor fit for small-batch LoRA training, as the minimum global batch size cannot be smaller than the world size.
Moreover, since LoRA fine-tuning is dominated by the cost of loading base model weights from HBM to SRAM \cite{lorafusion}, FSDP's redundant replication of LoRA adapter matrices across ranks further exacerbates the inefficiency.

To resolve this tension, we take two solutions.
First, on a single device, multiple LoRA adapters can be co-located on the same GPU, sharing the frozen backbone model and executing concurrently through grouped GEMM kernels.
Each adapter retains its own small per-adapter batch size, yet the aggregate workload fully saturates GPU memory and compute.
This \emph{batched multi-LoRA training} also synergizes with the early-exit strategy from observation~1: we run many configurations concurrently during the warmup phase, quickly identify the best candidates, evict the underperformers, and concentrate resources on the surviving configurations.

Second, for multi-device training, we introduce \emph{Adapter Parallelism} (AP), a parallelism strategy tailored to multi-LoRA workloads.
Like FSDP, AP shards the base model weights across GPU ranks; unlike FSDP, each rank is assigned a \emph{different} LoRA adapter rather than a different micro-batch of the same adapter.
This design offers three key advantages:
(i)~it supports per-adapter batch sizes as small as~1 without leaving any rank idle, since every rank always trains a distinct adapter;
(ii)~it eliminates redundant HBM-to-SRAM transfers of adapter parameters, as each adapter's low-rank matrices reside on exactly one rank; (iii)~it removes the gradient communication between different ranks for the gradients of LoRA adapters.
Together, batched multi-LoRA training and AP enable~{\NAME} to maintain high hardware utilization across single- and multi-GPU setups while respecting the small batch sizes that LoRA training demands.

\begin{figure}[t]
    \centering
    \includegraphics[width=0.95\columnwidth]{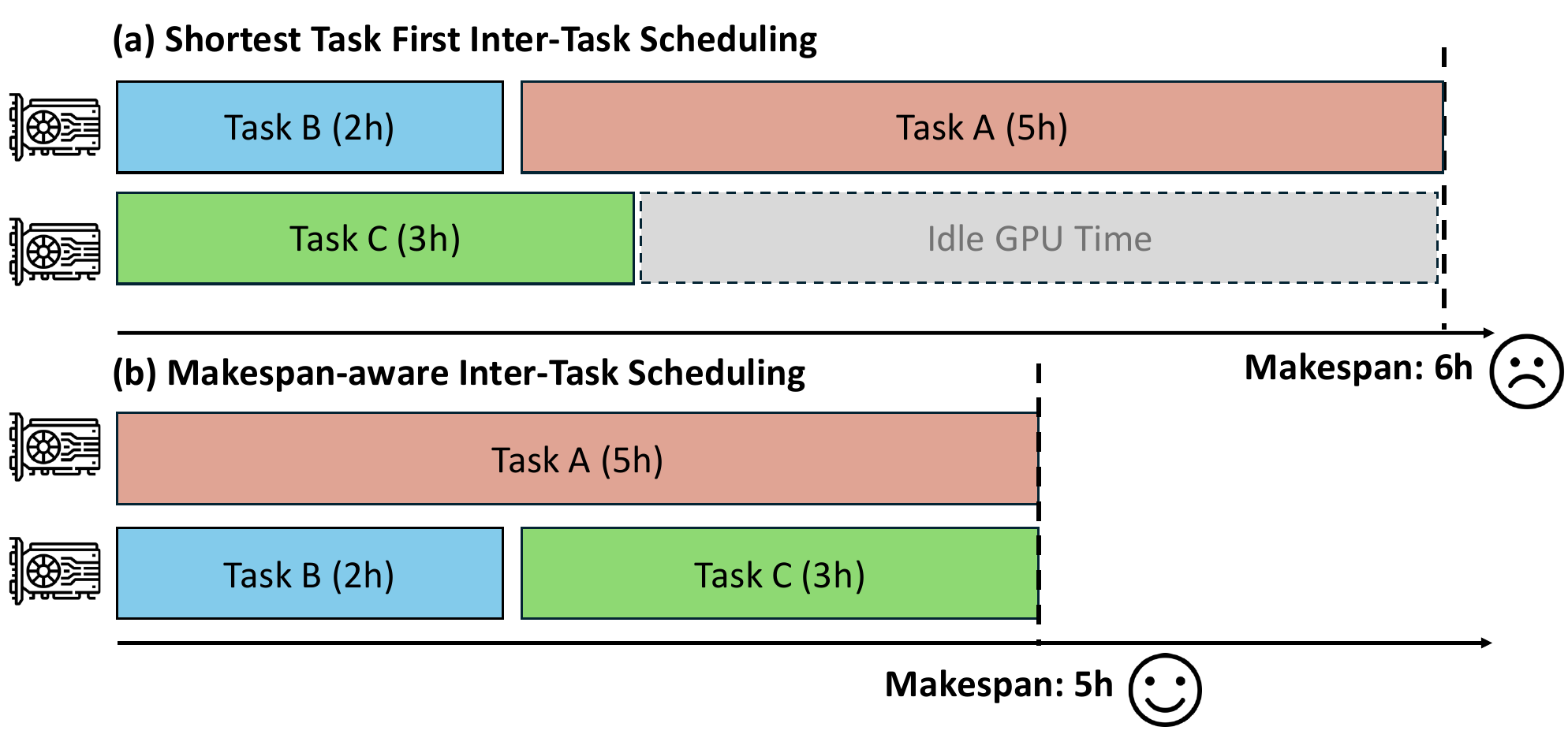}
    \vspace{-5pt}
    \caption{(a) Shortest-Job-First serve fails to fully utilize the GPU resources when multiple training tasks come. (b) Makespan-aware inter-task scheduling can fully minimize the overall training makespan.}
    \label{fig:scheduling_motivation}
    \vspace{-15pt}
\end{figure}

\textbf{Observation 3: Predictability of task-completion time creates optimization opportunities for workload scheduling.}
In a realistic training service, multiple LoRA training tasks from different users arrive together, each targeting a different base model, dataset, and hyperparameter search space.
These tasks differ in both resource footprint and duration.
A na\"ive shortest-task-first scheduler ignores both dimensions and can severely inflate makespan, for example by placing short tasks first and leaving GPUs idle while a long task runs alone (Figure~\ref{fig:scheduling_motivation}).
Such a problem would be further exacerbated when tasks have different GPU requirements, as poor placement can fragment the cluster and leave GPUs largely underutilized.

Nonetheless, unlike other machine learning tasks such as LLM serving workloads whose makespans are unpredictable, LoRA hyperparameter tuning tasks expose key scheduling information \emph{before} execution:
the number of configurations, the per-configuration step count, and the per-step training time together yield a reliable estimated duration, while the GPU requirement is determined by the base model size.
This predictability transforms scheduling from an online, reactive problem into an offline optimization: given tasks with known durations and GPU requirements, find an assignment to time slots and GPU partitions that minimizes overall makespan.
We formulate this as a variant of the heterogeneous-resource strip-packing problem and solve it efficiently in \S\ref{sec:scheduling}.
The scheduler also integrates with the early-exit mechanism from Observation~1: when a task terminates early due to convergence detection, freed GPUs are immediately reallocated to pending tasks in the entire system.

\section{{\NAME} System Overview}
\label{sec:design}


  To address the challenges of providing efficient LoRA training services we identified in \S\ref{sec:motivation}, we propose {\NAME}, a novel training system designed to efficiently handle heterogeneous LLM fine-tuning tasks. Unlike traditional frameworks that process tasks independently, {\NAME} jointly optimizes hyperparameter selection, hardware execution, and workload scheduling, as illustrated in Figure~\ref{fig:sys-overview}.

  This design is realized through three co-designed techniques.
  First, \emph{loss-aware early exit} (\S\ref{sec:early-exit}) detects diverging, overfitting, and underperforming configurations from their loss trajectories and terminates them before they consume their full training budget.
  Second, \emph{batched multi-LoRA execution} (\S\ref{sec:batched-engine}) co-locates multiple adapters on a shared frozen backbone via fused grouped GEMM kernels, and \emph{adapter parallelism} scales to multi-GPU settings by assigning each rank a distinct set of adapters rather than micro-batches of the same adapter.
  Third, \emph{hierarchical scheduling} (\S\ref{sec:scheduling}) coordinates the dynamic workload at two levels: intra-task scheduling packs adapters onto executors to maximize GPU utilization, while inter-task scheduling minimizes the total makespan of all tasks via optimization solving.

  These three techniques together enable {\NAME} to operate as a \emph{LoRA-as-a-Service}: rather than requiring users to manually launch, monitor, and tune individual training jobs, the system accepts a declarative task specification (a base model, dataset, and hyperparameter search space, as shown in Listing~\ref{lst:alto}) and autonomously returns the best-performing adapter.
  Internally, {\NAME} profiles each task's throughput and memory footprint, computes a task placement plan, and instantiates one \emph{executor} per task that hosts multiple jobs concurrently on a shared base-model replica.
  Throughout training, the system continuously monitors loss trajectories, switches training jobs, and replans task placement upon task completion, all transparently to the user.


\section{Loss-Aware Early Exit}
\label{sec:early-exit}

\begin{figure}
    \centering
    \includegraphics[width=1\linewidth]{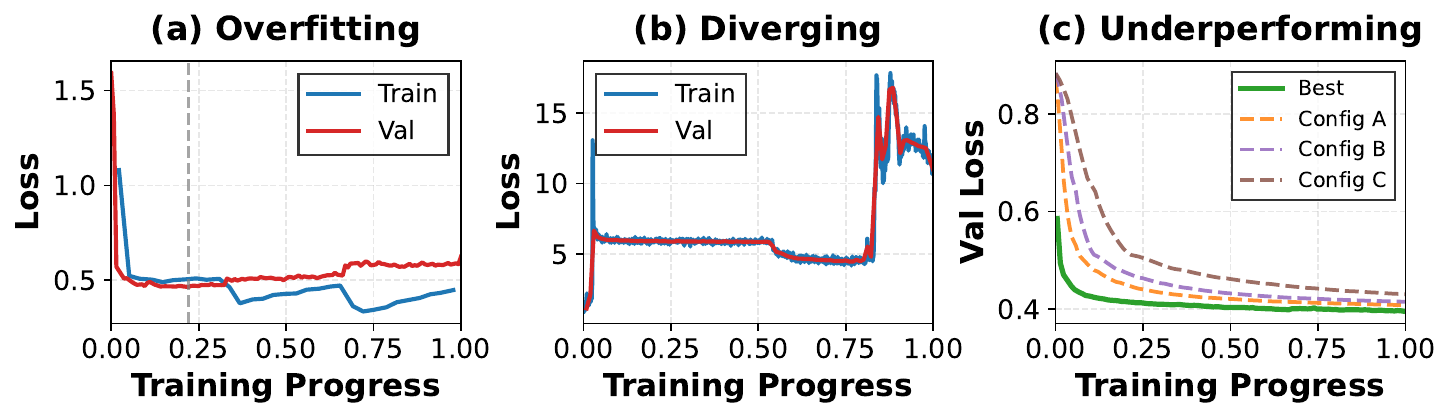}
    \caption{The illustration of the loss curves of three typical redundant training patterns: (a) Overfitting, (b) Diverging, and (c) Underperforming.}
    \vspace{-10pt}
    \label{fig:redundant_training_patterns}
\end{figure}

To exploit detectable redundancy during training, {\NAME} divides each task into a \textit{warmup stage} and a \textit{continue-training stage}. After warmup, only the top-performing adapters advance to continue-training, where online pattern detection continuously monitors loss trajectories and terminates diverging or overfitting jobs.

\subsection{Online Pattern-Based Exiting}
\label{sec:pattern-detection}

As motivated in \S\ref{sec:motiv_early_exit}, divergence and overfitting are detectable online from training and validation loss curves. We smooth training loss with an exponential moving average (EMA): $\hat\ell_t = \alpha \cdot \ell_t + (1 - \alpha) \cdot \hat\ell_{t-1}$, and use raw validation loss directly. Algorithm~\ref{alg:pattern-detection} gives the unified procedure; underperformance filtering, which requires cross-adapter comparison, is handled at the warmup boundary (\S\ref{sec:warmup}).

\begin{figure}[t]
    \centering
    \includegraphics[width=0.99\columnwidth]{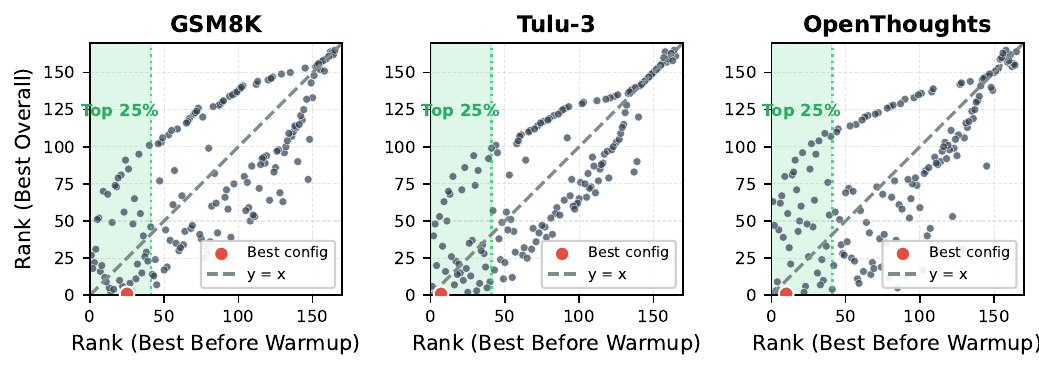}
    \caption{Rank correlation between validation loss at the end of the warmup phase and at the end of training. The relative ordering of well-behaved configurations is largely preserved, and the eventually best configuration is always preserved in top 25\% after the warmup, validating the effectiveness of warmup-based exiting.}
    \vspace{-10pt}
    \label{fig:monotonicity}
\end{figure}

\vspace{1pt}
\noindent\textbf{Pattern-1: Divergence.}
A hyperparameter configuration may cause the model to diverge at any point during training, typically due to a learning rate that is too high, a LoRA rank mismatched with the task complexity, or a learning rate schedule that interacts poorly with the data distribution. When divergence occurs, both the training loss and the validation loss trend upward simultaneously (Figure~\ref{fig:redundant_training_patterns}b). We detect this by fitting a linear regression over the most recent $w$ loss values independently. If both slopes exceed a positive threshold $\tau_{\text{slope}}$ for $p_{\text{div}}$ consecutive evaluation steps, the adapter is terminated immediately. The patience counter resets to zero whenever either slope drops below $\tau_{\text{slope}}$, preventing spurious exits from transient gradient spikes. 

\vspace{1pt}
\noindent\textbf{Pattern-2: Overfitting.}
When a configuration is trained for too long or with an overly aggressive learning rate, the training loss continues to decrease while the validation loss turns upward, leading to overfitting (Figure~\ref{fig:redundant_training_patterns}a). This is common in LoRA fine-tuning with multi-epoch schedules, where the small adapter capacity memorizes the training distribution without generalizing. We detect overfitting by monitoring the gap ratio between the raw validation loss and the EMA-smoothed training loss: $g = (\ell_{\text{val}} - \hat\ell_{\text{train}}) / \hat\ell_{\text{train}}$. When $g$ exceeds a threshold $\tau_{\text{gap}}$ for $p_{\text{ovf}}$ consecutive evaluation steps, the adapter is checkpointed at its best validation loss and then terminated. Transient validation fluctuations reset the patience counter, so only a sustained gap triggers an exit. By checkpointing the best model before termination, this mechanism recovers the optimal adapter without any post-hoc checkpoint selection.

\begin{algorithm}[t]
\caption{Loss-Aware Pattern Detection.}
\label{alg:pattern-detection}
\small
\begin{algorithmic}[1]
\REQUIRE EMA train losses $\hat\ell_{\text{train}}$, raw val losses $\ell_{\text{val}}$, window $w$, thresholds $\tau_{\text{slope}}$, $\tau_{\text{gap}}$, patience $p_{\text{div}}$, $p_{\text{ovf}}$
\ENSURE Set of exit decisions for each adapter
\item[\commentone{Pattern 1: Divergence detection.}]
\IF{$|\hat\ell_{\text{train}}| \geq w$ \AND $|\ell_{\text{val}}| \geq w$}
    \STATE{$s_{\text{train}} \leftarrow \textbf{linregSlope}(\hat\ell_{\text{train}}[-w:])$}
    \STATE{$s_{\text{val}} \leftarrow \textbf{linregSlope}(\ell_{\text{val}}[-w:])$}
    \STATE{$\mathit{cnt}_{\text{div}} \leftarrow \mathbf{if}\; s_{\text{train}} \!\geq\! \tau_{\text{slope}} \wedge s_{\text{val}} \!\geq\! \tau_{\text{slope}} \;\mathbf{then}\; \mathit{cnt}_{\text{div}}\!+\!1 \;\mathbf{else}\; 0$}
    \IF{$\mathit{cnt}_{\text{div}} \geq p_{\text{div}}$} \RETURN \textsc{Exit}(reason=\texttt{diverging}) \ENDIF
\ENDIF
\item[\commentone{Pattern 2: Overfitting detection.}]
\STATE{$g \leftarrow (\ell_{\text{val}}[-1] - \hat\ell_{\text{train}}[-1]) / \hat\ell_{\text{train}}[-1]$}
\STATE{$\mathit{cnt}_{\text{ovf}} \leftarrow \mathbf{if}\; g > \tau_{\text{gap}} \;\mathbf{then}\; \mathit{cnt}_{\text{ovf}}\!+\!1 \;\mathbf{else}\; 0$}
\IF{$\mathit{cnt}_{\text{ovf}} \geq p_{\text{ovf}}$}
    \STATE{\textbf{checkpoint}(best val.\ loss model);\quad} \RETURN \textsc{Exit}(reason=\texttt{overfitting})
\ENDIF
\item[\commentone{Pattern 3: Underperformance (warmup boundary only).}]
\IF{warmup finished}
    \STATE{$\mathit{ranked} \leftarrow \textbf{sortByValLoss}(\text{all surviving adapters})$}
    \STATE{$k \leftarrow \lceil \mathit{warmup\_select\_ratio} \times |\mathit{ranked}| \rceil$}
    \FOR{each adapter in $\mathit{ranked}[k:]$}
        \RETURN \textsc{Exit}(reason=\texttt{underperforming})
    \ENDFOR
\ENDIF
\end{algorithmic}
\end{algorithm}


\subsection{Warmup-Based Exiting}
\label{sec:warmup}

Given $K$ candidate configurations for a task, {\NAME} runs each through a brief warmup phase before committing to full training. Online pattern detection is already active during warmup, terminating clearly diverging configurations and freeing their slots for queued candidates to rotate in. This allows the system to cycle through all $K$ candidates even when only a fraction can run concurrently on a single GPU.

At the warmup boundary, each surviving adapter is evaluated on the validation set. The scheduler ranks them by validation loss and retains the top-$k$ candidates, evicting the rest—persistently bottom-ranked configurations are unlikely to recover (Figure~\ref{fig:redundant_training_patterns}c). Evicted adapters' parameters and optimizer states are discarded to reclaim GPU memory, while selected adapters transition into continue-training carrying over their optimizer states and loss histories.
Our rank-correlation analysis (Figure~\ref{fig:monotonicity}) confirms that relative configuration ordering after warmup strongly correlates with final performance, validating this selection. In practice, a warmup of 5\% of total steps with a 25\% retention ratio provides an optimal balance (Appendix~\ref{sec:appendix-sensitivity}).
\section{Dynamically Batched Executor}
\label{sec:batched-engine}

This section presents two complementary techniques to co-locate multiple adapters efficiently: \emph{grouped GEMM execution} for single-GPU batching (\S\ref{sec:grouped_gemm}), and rank-local \emph{adapter parallelism} for multi-GPU scaling (\S\ref{sec:adapter-parallelism}).

\subsection{Decoupled Base-LoRA Execution}
\label{sec:grouped_gemm}

All LoRA adapters share the same frozen base model, so a single forward pass through the backbone can serve multiple adapters simultaneously.
The key observation is that the base GEMM ($Y = XW$) is \emph{compute-bound} on large weight matrices, while the LoRA GEMMs ($S = XA$, $L = SB$) are \emph{memory-bandwidth bound} due to low rank ($r \ll H$)~\cite{lorafusion}.
This asymmetry is central to our design.

Existing systems make suboptimal tradeoffs along this boundary.
mLoRA~\cite{ye2024mlora} preserves cuBLAS for the base GEMM but executes the LoRA path with $3N$ separate kernel launches per layer, incurring launch overhead and poor SM utilization.
LoRAFusion~\cite{lorafusion} fuses base and LoRA into one Triton kernel, but sacrifices cuBLAS throughput (10--20\%~\cite{tillet2019triton}) and introduces FLOP waste: its wide-GEMM formulation executes $(\sum_i L_i)(\sum_i r_i)$ FLOPs while only $\sum_i L_i r_i$ are useful.

\vspace{1pt}
\noindent\textbf{Decoupled grouped GEMM.}
{\NAME} avoids both pitfalls by treating the two paths independently.
The base GEMM runs via cuBLAS on the concatenated batch; the LoRA path uses two Triton kernels: a \emph{grouped GEMM} computing $\{S_i = X_i A_i\}$ in a single launch, and a \emph{fused GEMM-add} computing $Y{=}S_i B_i + Y_{\text{base}}$ directly.
This reduces LoRA launches from $O(N)$ to $O(1)$ per layer, concatenating thread blocks across adapters for full SM occupancy, while computing only the diagonal blocks $S_i{=}X_i A_i$ with zero wasted FLOPs.

\vspace{1pt}
\noindent\textbf{Backward pass.}
Training backward passes are comparably expensive to forward, yet the serving kernels above lack backward support entirely.
{\NAME} computes input gradients ($dS{=}\text{scale}{\cdot}dY\, B_i^\top$, $dX{=}dS\, A_i^\top$) in a single grouped Triton kernel reusing the same $O(1)$-launch schedule, while weight gradients ($dA_i{=}X_i^\top dS_i$, $dB_i{=}S_i^\top dY_i$) are batched via \texttt{grouped\_mm}, also in $O(1)$ launches.
The forward caches intermediate $S$ to avoid recomputation, trading modest memory for a saved kernel launch per layer.

\subsection{Rank-Local Adapter Parallelism}
\label{sec:adapter-parallelism}

\begin{figure}[t]
  \centering
  \includegraphics[width=0.99\columnwidth]{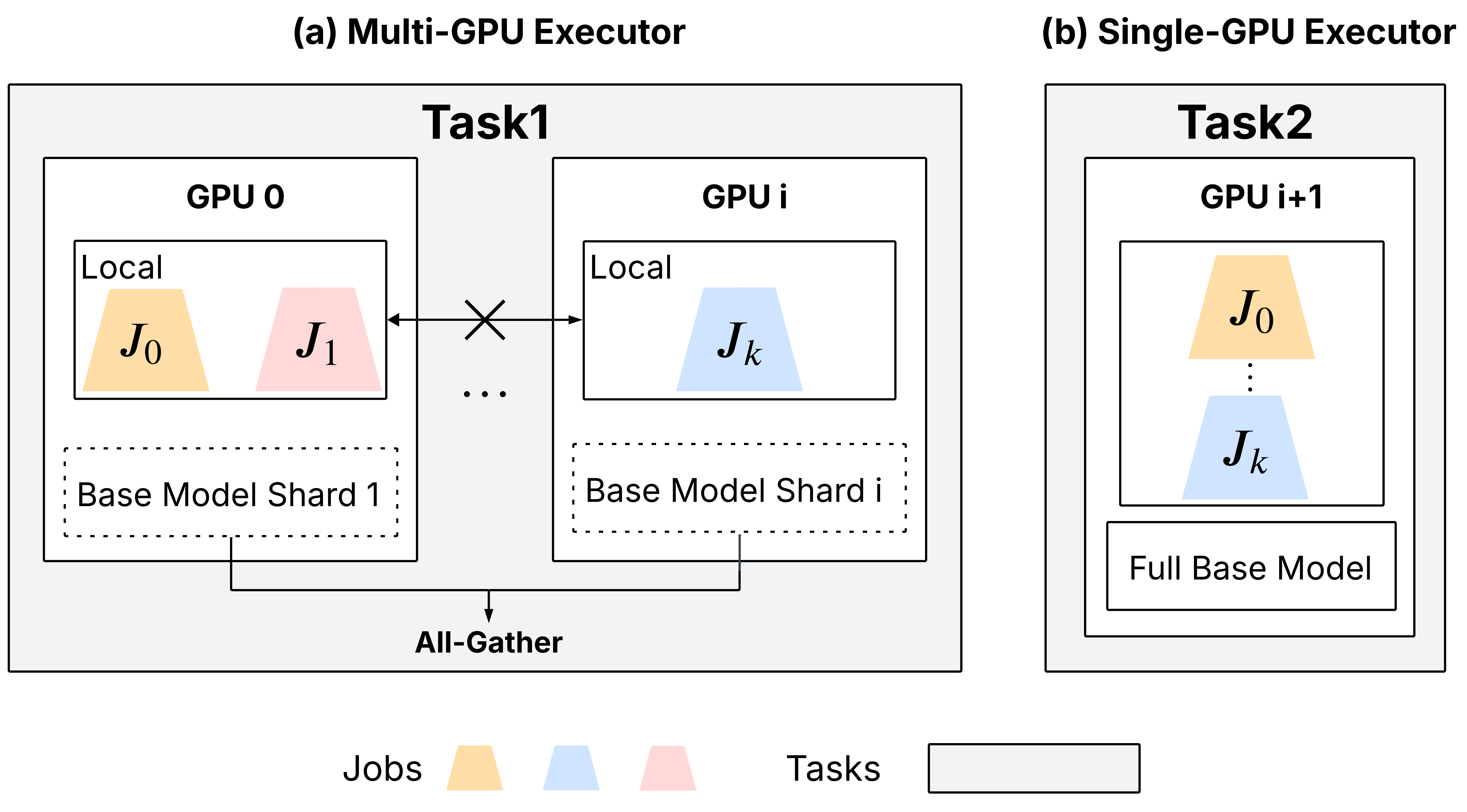}
  \caption{Two executor modes in \NAME's batched execution engine.
(a)~\emph{Multi-GPU executor} with adapter parallelism: base model weights are sharded across ranks and synchronized via all-gather, while each rank trains a disjoint set of LoRA jobs locally via grouped GEMM, therefore there is no adapter gradients cross rank boundaries.
(b)~\emph{Single-GPU executor}: the full base model and multiple LoRA jobs reside on a single device, trained concurrently through grouped GEMM batching.}
  \label{fig:adapter-parallelism}
  \vspace{-20pt}
\end{figure}

When the base model exceeds single-GPU memory, we shard it across ranks using FSDP~\cite{zhao2023pytorch}.
However, instead of distributing micro-batches of the same adapter across ranks, we propose \emph{adapter parallelism}, which distributes different adapters to different ranks (as shown in Figure~\ref{fig:adapter-parallelism}).
Each rank maintains a disjoint set of LoRA adapters with independent parameters, optimizers, and loss trackers.
During forward and backward passes, all ranks participate in FSDP all-gather for base model weights, but LoRA computation and gradients remain purely local, which incurs no cross-rank communication of adapter parameters.

This design offers three key advantages over standard FSDP for LoRA workloads.
First, \emph{no rank is ever idle}: each rank trains its own adapters with full mini-batches regardless of per-adapter batch size, whereas FSDP would leave $P{-}1$ ranks idle when the global batch size is 1.
Second, \emph{adapter gradient communication is eliminated}: each adapter's low-rank matrices reside on exactly one rank and are never replicated or synchronized, removing the gradient all-reduce that FSDP would impose on adapter parameters.
Third, \emph{redundant adapter weight transfers are removed}: in standard FSDP, every rank replicates the same adapter and independently loads its weights from HBM to SRAM during each forward and backward pass, incurring $P\times$ redundant memory traffic on the bandwidth-bound LoRA path.
With adapter parallelism, each adapter is read from HBM on exactly one rank, eliminating this redundancy entirely.
Adapter parallelism composes naturally with grouped GEMM: within each rank, multiple adapters can be co-located, compounding single-GPU and multi-GPU efficiency gains.

\section{Hierarchical Scheduling}
\label{sec:scheduling}

The batched multi-LoRA engine and loss-aware early exit create a dynamic workload where jobs are continuously admitted/terminated. {\NAME} manages this dynamics with a hierarchical scheduler: an \emph{online intra-task scheduler} that performs admission and backfill within each executor (\S\ref{sec:intra-task}), and a \emph{dynamic inter-task scheduler} that places tasks across GPUs and time slots to minimize cluster makespan (\S\ref{sec:inter-task}).

\vspace{-0.5em}
\subsection{Online Greedy Intra-Task Scheduling}
\label{sec:intra-task}

Within a single task, the intra-task scheduler must continuously decide how many adapters to co-locate on each executor and when to admit/evict them. These decisions depend on the GPU memory and the throughput impact of adapter packing density, both of which vary with the base model, LoRA rank, sequence length, and per-adapter batch size.

\label{sec:mem-profiling}
Rather than requiring the user to specify packing limits manually, {\NAME} runs an automatic memory profiling stage before training begins, fitting a lightweight linear model $\hat{M}(B)$ that predicts peak HBM usage as a function of total batch size $B$ (Appendix~\ref{sec:appendix-intra-task}). At runtime, the scheduler groups adapters by per-adapter batch size to maximize grouped GEMM efficiency and greedily admits them in decreasing batch-size order, admitting an adapter only if $\hat{M}$ confirms the new configuration fits within a safety margin. When an adapter exits, the scheduler backfills the vacated slot, which prefers a pending job of the same batch size to preserve homogeneous packing, but accepts mixed configurations when necessary.


\subsection{Dynamic Inter-Task Scheduling}
\label{sec:inter-task}

When a user submits multiple tasks, e.g., fine-tuning Llama-8B on GSM8K, the inter-task scheduler must partition $G$ GPUs across tasks over time to minimize end-to-end makespan. Unlike LLM serving requests whose runtimes are unpredictable, LoRA fine-tuning tasks expose reliable duration estimates before execution. Because our scheduling constraints are highly structured, we can compute optimal makespan schedules in under a second. This negligible overhead enables {\NAME} to operate as a fully dynamic online scheduler rather than relying on static offline plans.

Before a task is scheduled, {\NAME} runs a lightweight throughput profiling phase. It launches a short training run to measure the throughput in samples per second. Combined with the total sample count, this yields an estimated task duration $d_i = \text{total\_samples}_i \;/\; \text{throughput}_i$. The GPU requirement $g_i$ for each task is determined by the base model size. Profiling results are cached per task to avoid redundant measurements.
%
Given $n$ tasks with profiled durations $d_i$ and GPU requirements $g_i$ on $G$ total GPUs, the problem is $P \mid \text{size}_j \mid C_{\max}$. We formulate it as a constraint programming problem. Table~\ref{tab:milp-vars} lists the decision variables; $M{=}\sum_i d_i$ is the big-M constant.

\begin{table}[t]
  \centering\small
 \caption{Constraint programming decision variables for inter-task scheduling.}
\label{tab:milp-vars}
\vspace{-5pt}
\scalebox{0.94}{
  \begin{tabular}{@{}ccl@{}}
    \toprule
    Variable & Domain & Description \\
    \midrule
    $s_i$ & $\mathbb{R}_{\geq 0}$ & Start time of task $i$ \\
    $x_{ig}$ & $\{0,1\}$ & 1 iff task $i$ occupies GPU $g$ \\
    $y_{ij}$ & $\{0,1\}$ & 1 iff task $i$ precedes $j$\; ($i{<}j$) \\
    $C_{\max}$ & $\mathbb{R}_{\geq 0}$ & Makespan \\
    \bottomrule
  \end{tabular}}
  \vspace{-10pt}
\end{table}

\vspace{-0.5em}

\vspace{-6pt}
\begin{equation*}
\begin{aligned}
\min \quad & C_{\max} \\
\text{s.t.} \quad & \textstyle\sum_{g}^G x_{ig} = g_i, \quad s_i + d_i \leq C_{\max} && \forall i \\
& \begin{cases} 
s_i + d_i \leq s_j + M(3 - x_{ig} - x_{jg} - y_{ij}) \\
s_j + d_j \leq s_i + M(2 - x_{ig} - x_{jg} + y_{ij})
\end{cases} && \forall i < j, g
\label{eq:schedule}
\end{aligned}
\end{equation*}

The above equations show big-M disjunctive no-overlap constraints: they activate only when tasks $i,j$ share GPU $g$ ($x_{ig}{=}x_{jg}{=}1$) and enforce temporal ordering via $y_{ij}$. A single $y_{ij}$ per pair suffices because each task occupies all its assigned GPUs simultaneously. The CP-SAT solver~\cite{cpsatlp} finds the optimum in ${<}\,1$\,s for all tested instances.

\noindent\textbf{Event-driven replanning.}
Unlike static schedulers, {\NAME} maintains a dynamic "living" queue. The inter-task scheduler sits on the critical path and replans the cluster assignment on the fly whenever a major state change occurs. Specifically, the inter-task scheduling is triggered by two events: (1) \emph{Task Arrival}: a new workload is submitted by a user, and (2) \emph{Task Completion}: a task successfully finishes its run (which frequently occurs earlier than the worst-case duration $d_i$ due to massive early-exits executed by the intra-task pattern detector (\S\ref{sec:pattern-detection})). This event-driven loop guarantees true multi-tenancy and zero resource waste, as freed GPUs are instantly backfilled with the next optimal task.

\section{Evaluation}
\label{sec:evaluation}
\begin{figure*}[t]
      \centering
      \includegraphics[width=\linewidth]{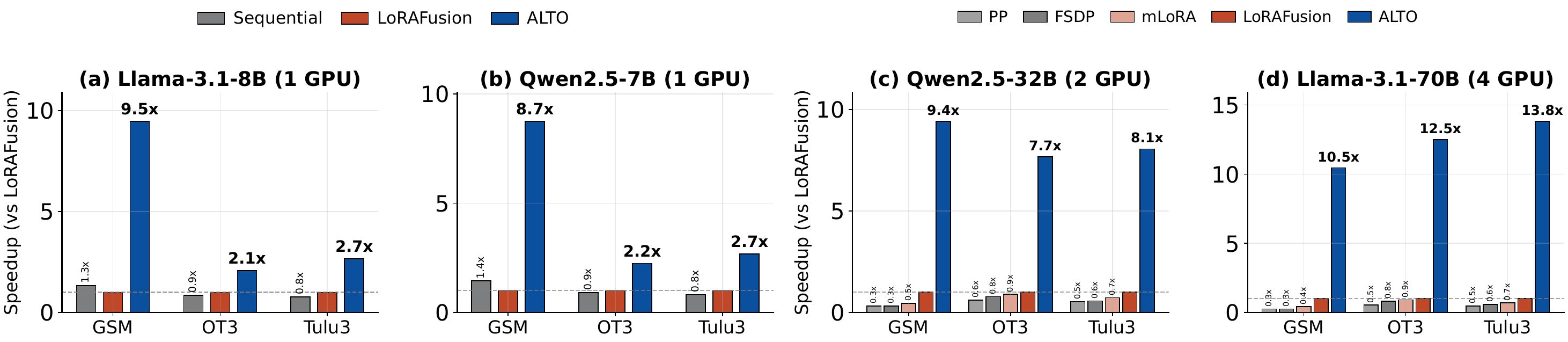}
      \caption{End-to-end training speedup of {\NAME} across single-GPU and multi-GPU configurations. From left to right: Llama-3.1-8B and Qwen2.5-7B on a single H100 GPU, Qwen2.5-32B on 2$\times$ H100 and Llama-3.1-70B on 4$\times$ H100. Each configuration trains 60 (single-GPU) or 64 (multi-GPU) heterogeneous LoRA adapters with varied ranks, batch sizes, and learning rates across three datasets. {\NAME} achieves up to 9.5$\times$ speedup on single-GPU and up to 13.8$\times$ speedup on multi-GPU settings, consistently outperforming all baselines. Speedup is evaluated relative to LoRAFusion.}
      \label{fig:e2e_speedup}
\end{figure*}

We evaluate {\NAME} on four LLMs spanning 7B to 70B parameters, three SFT datasets, one DPO dataset, and both single-GPU and multi-GPU configurations. Our evaluation answers the following three major questions:
\begin{itemize}[leftmargin=*,nosep]
  \item How does {\NAME} compare end-to-end against sequential training and state-of-the-art multi-LoRA baselines? (\S\ref{sec:e2e})
  \item Does batched execution with early exit preserve final adapter quality? (\S\ref{sec:e2e})
  \item How much does each component (e.g., batching, early exit, and scheduling) contribute to the overall speedup? (\S\ref{sec:ablation})
\end{itemize}

\subsection{Experimental Setup}
\label{sec:eval-setup}

\textbf{Hardware.} All experiments are conducted on NVIDIA H100 SXM5 80GB GPUs interconnected via NVLink.
Single-GPU experiments use one H100, while multi-GPU experiments use 2$\times$H100 for Qwen2.5-32B and 4$\times$H100 for Llama-3.1-70B.

\noindent \textbf{Models.} 
We evaluate {\NAME} on four widely-used open-source LLMs spanning different scales:
Llama-3.1-8B~\cite{grattafiori2024llama} and Qwen2.5-7B~\cite{qwen2025qwen25technicalreport} for single-GPU experiments,
and Qwen2.5-32B and Llama-3.1-70B for multi-GPU experiments with Adapter Parallelism.

\noindent \textbf{Datasets.}
We use three diverse datasets for supervised fine-tuning (SFT):
(1)~{GSM8K}~\cite{cobbe2021training} (GSM), a grade-school math reasoning dataset where we use 90\% of the training set for training and 10\% for validation, and the full test set for evaluation;
(2)~{Tulu-3}~\cite{lambert2024tulu}, an instruction-following dataset using a subset of 9,000 training samples, 1,000 for validation, and 1,000 for evaluation; and
(3)~{OpenThoughts3}~\cite{guha2506openthoughts} (OT3), a general reasoning dataset using the same 9,000/1,000/1,000 split.
The maximum sequence length is 1,024 tokens for GSM8K and Tulu-3, and 2,048 tokens for OpenThoughts3.
For direct preference optimization (DPO) experiments, we use UltraFeedback~\cite{cui2023ultrafeedback} for evaluation.

\noindent \textbf{Hyperparameter search space.}
We search over 60 (single-GPU) or 64 (multi-GPU) configurations spanning learning rates 1e-5--5e-4, LoRA ranks 16--128, and per-adapter batch sizes 1--8.
All runs use 3 epochs, paged AdamW 8-bit, and LoRA on all attention and MLP projections with $\alpha = 2r$.
Full details are in Appendix~\ref{sec:appendix-training-details}.

\vspace{2pt}
\noindent \textbf{Baselines.}
We compare against:
(1)~{Sequential}, which trains each LoRA adapter one at a time on a single GPU;
(2)~{LoRAFusion}~\cite{lorafusion}, a state-of-the-art batched multi-LoRA system with fused GEMM kernels;
(3)~{mLoRA}~\cite{ye2024mlora}, a multi-LoRA training framework supporting concurrent adapter training.
For multi-GPU experiments, we additionally compare against
(4)~{Pipeline Parallelism}, which partitions the model across GPUs and processes adapters sequentially.

\vspace{2pt}
\noindent \textbf{Metrics.}
We report end-to-end training speedup (wall-clock time to complete all adapter training) normalized to the sequential baseline for single-GPU and pipeline parallelism for multi-GPU settings.
For GSM8K, we evaluate downstream accuracy using vLLM~\cite{kwon2023efficient} with zero-shot prompting and strict answer parsing, selecting the checkpoint with the lowest validation loss for each configuration.
For Tulu-3 and OpenThoughts3, we report validation loss as these datasets lack ground-truth answers for automated evaluation.

\vspace{2pt}
\noindent \textbf{Implementation.}
{\NAME} is built on PyTorch 2.9.1 / CUDA 12.1 with Triton 3.5.1 for the fused grouped GEMM kernels; evaluation inference uses vLLM 0.14.1.

\subsection{End-to-End Performance}
\label{sec:e2e}

\textbf{Single-GPU speedup.}
Figure~\ref{fig:e2e_speedup} (a-b) reports end-to-end wall-clock speedup on Llama-3.1-8B and Qwen2.5-7B, each training 60 heterogeneous LoRA adapters across three datasets.
{\NAME} achieves up to $9.5\times$ speedup over the LoRAFusion baseline, consistently outperforming all baselines across all six single-GPU model--dataset combinations.

\noindent \textbf{Multi-GPU speedup.}
Figure~\ref{fig:e2e_speedup} (c-d) shows the results on Qwen2.5-32B (2$\times$H100) and Llama-3.1-70B (4$\times$H100), each training 64 adapters across three datasets.
{\NAME} achieves up to $13.8\times$ speedup over the LoRAFusion baseline.
The larger gains stem from Adapter Parallelism (\S\ref{sec:adapter-parallelism}), which assigns each GPU rank a distinct set of adapters, whereas pipeline parallelism leaves most stages idle when per-adapter batch sizes are small.
This advantage is consistent across both the 32B (2-GPU) and 70B (4-GPU) scales, confirming that AP's communication efficiency generalizes as the model and GPU count grow.

\begin{figure*}[t]
    \centering
     \includegraphics[width=\linewidth]{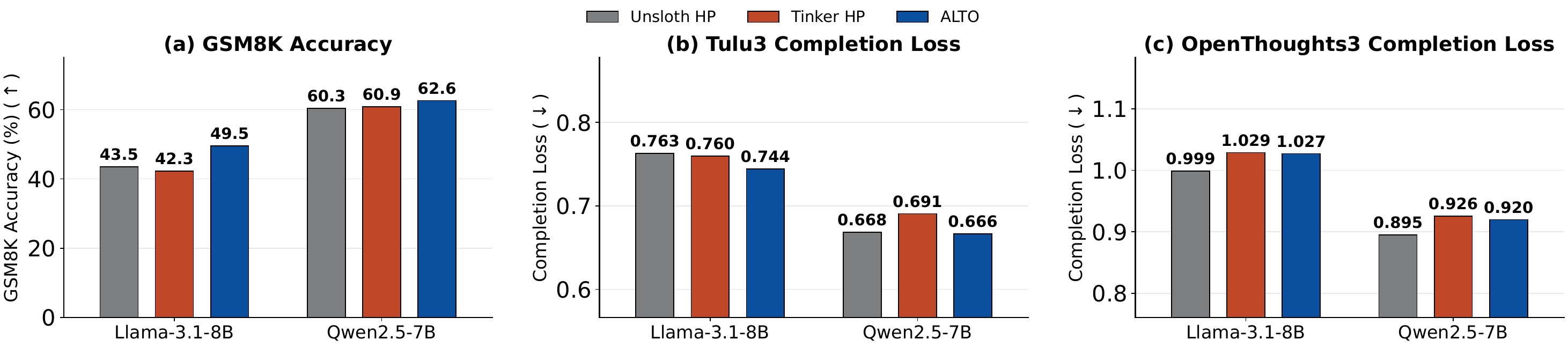}
    \caption{Model quality of the best configuration found by {\NAME} on single-GPU 7B--8B models, compared against expert-recommended hyperparameters from Unsloth and Tinker. (a)~GSM8K test accuracy (higher is better). Completion loss on the evaluation split of (b) Tulu-3 and (c) OpenThoughts3 (lower is better). For {\NAME}, each bar shows the best result found across batched runs with per-adapter batch sizes in $\{1,2,4,8\}$.}
    \label{fig:combined-accuracy}
\end{figure*}

 \noindent \textbf{Quality preservation.}
Beyond throughput, we verify that {\NAME} preserves final adapter quality despite batched execution and early termination of unpromising configurations.
Figure~\ref{fig:combined-accuracy} compares the best configuration found by {\NAME} against expert-recommended hyperparameters from Unsloth and Tinker on GSM8K test accuracy (a) and completion loss on the evaluation set of Tulu-3 (b) and OpenThoughts3 (c).
{\NAME} matches or exceeds the expert-recommended settings across all model--dataset combinations.
This result validates that the early-exit mechanism correctly preserves the best-performing configurations while eliminating redundant training.
Notably, expert-recommended hyperparameters often fail to achieve the best results, underscoring the necessity of systematic hyperparameter search.

\begin{figure}[t]
    \centering
    \includegraphics[width=0.72\linewidth]{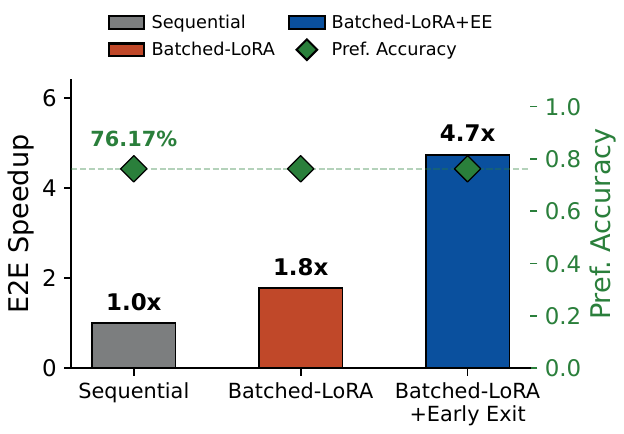}
    \caption{DPO training on UltraFeedback. Bars show end-to-end speedup over sequential training; diamond markers show best preference accuracy across all per-adapter batch sizes. {\NAME} with early exit achieves $4.7\times$ speedup while preserving the same preference accuracy (76.2\%) as Batched-LoRA without early exit.}
    \label{fig:rl-dpo-e2e}
\end{figure}

\noindent \textbf{RL End-to-end results.}
We extend {\NAME} to reinforcement learning from human feedback (RLHF) by evaluating on DPO~\cite{rafailov2023direct} with UltraFeedback~\cite{cui2023ultrafeedback}.
Figure~\ref{fig:rl-dpo-e2e} reports the end-to-end speedup and best preference accuracy across 60 configurations with per-adapter batch sizes $\in \{2,4,8,16\}$.
{\NAME} with early exit achieves $4.7\times$ speedup over sequential training and $2.7\times$ over Batched-LoRA alone, while the best configuration found by early exit attains the same $76.2\%$ preference accuracy as Batched-LoRA without early exit, confirming that early termination does not discard the top-performing adapter.

\begin{figure}[t]
    \centering
    \includegraphics[width=0.9\linewidth]{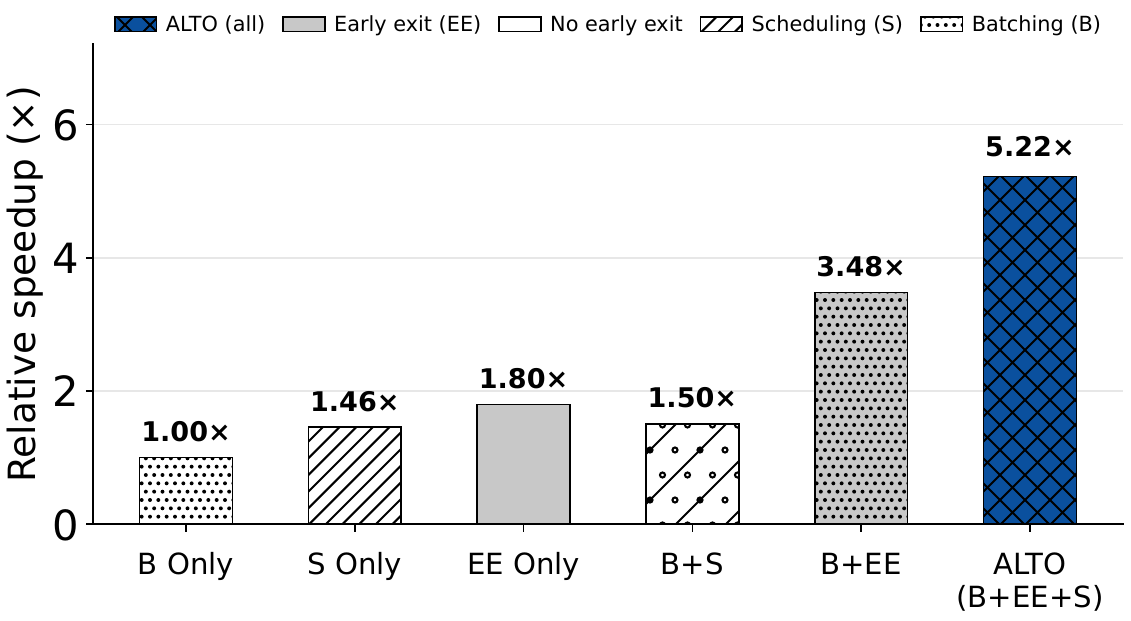}
    \caption{Evalution of {\NAME} components on 8-GPU training makespan. B\,=\,Batched LoRA, S\,=\,Scheduler, EE\,=\,Early Exit. The full system (B+S+EE) achieves a $5.2\times$ reduction in makespan compared to batching alone (B), with early exit contributing the largest individual gain.}
    \label{fig:ablation_makespan}
\end{figure}

\noindent \textbf{Inter-task scheduling.}
We evaluate the inter-task scheduler on 8$\times$H100 GPUs with 11 heterogeneous tasks spanning four model scales: Llama-3.1-70B (4 GPUs per task), Qwen2.5-32B (2 GPUs), and Llama-3.1-8B / Qwen2.5-7B (1 GPU each), each training adapters on GSM8K at varied batch sizes. This setting exercises the scheduler's ability to bin-pack tasks with diverse GPU requirements onto a shared cluster.
Figure~\ref{fig:ablation_makespan} ablates the three components on overall makespan: Batched LoRA (B), Scheduler (S), and Early Exit (EE).
Adding early exit (B+EE) yields the largest incremental gain by eliminating wasted training and freeing GPU slots for remaining tasks.
The full system (B+S+EE) achieves a $5.2\times$ reduction in makespan compared to batching alone, demonstrating that the three components are complementary: early exit reduces per-task compute, the scheduler reduces cross-task fragmentation, and batching amortizes per-step cost.

\subsection{Ablation Studies}
\label{sec:ablation}
We isolate the contribution of each component in {\NAME} through ablation experiments on kernel performance, quality impact of early exit, and training sample savings.

\noindent \textbf{Kernel contributions.}
Table~\ref{tab:kernel-micro} compares the training time of {\NAME}'s fused grouped Triton kernels against PyTorch back-to-back execution (per-adapter LoRA kernels run sequentially on the mixed batch) and fully sequential training (each adapter trained alone, wall-clock times summed).
The fused kernels achieve $1.36$--$1.91\times$ speedup over PyTorch and $2.5$--$5.1\times$ speedup over sequential training.
Gains scale inversely with per-adapter batch size ($1.91\times$ at BS=1 vs.\ $1.36\times$ at BS=4), consistent with the LoRA path's increasing share of total compute at smaller batches (\S\ref{sec:batched-engine}).


\begin{figure}[t]
    \centering \includegraphics[width=\linewidth]{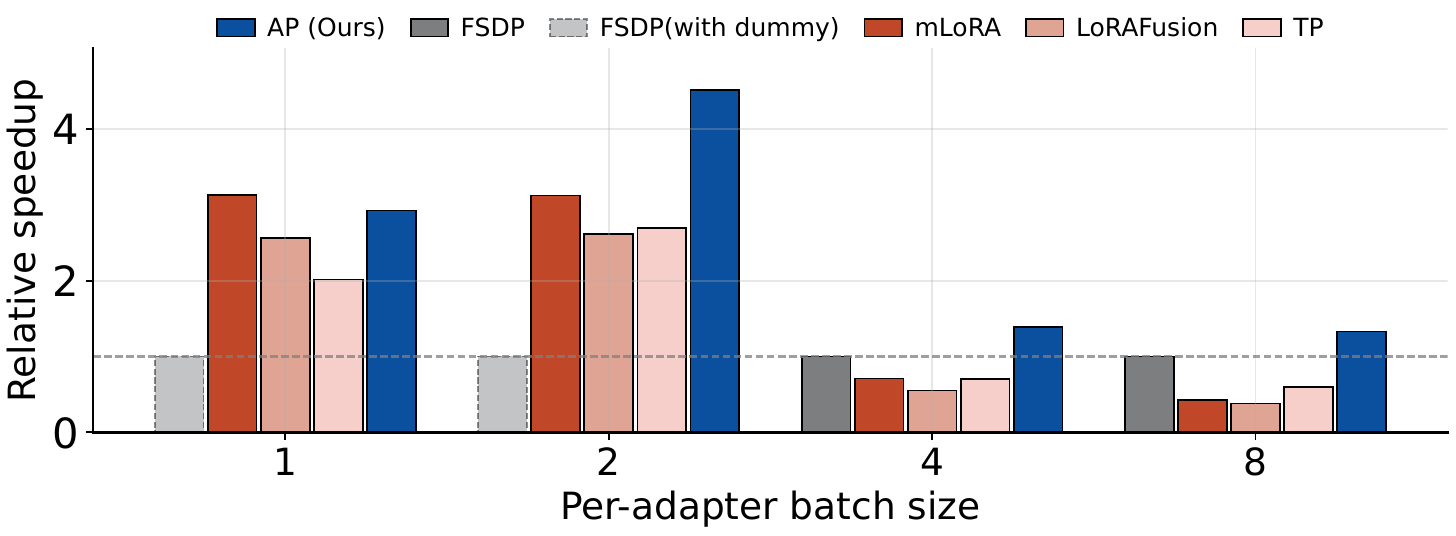}
    \caption{Adapter Parallelism (AP) microbenchmark on 4$\times$H100 GPUs: speedup relative to FSDP across per-adapter batch sizes, with 8 adapters and sequence length 256. AP (Ours) outperforms FSDP, TP, mLoRA, and LoRAFusion at all batch sizes, peaking at $4.7\times$ at per-adapter batch size~2.}
    \label{fig:ap-speedup}
\end{figure}

\vspace{1pt}
\noindent \textbf{Adapter Parallelism.}
Figure~\ref{fig:ap-speedup} microbenchmarks Adapter Parallelism (AP) against FSDP, TP, and mLoRA on 4$\times$H100 GPUs, training 8 adapters with sequence length 256 across per-adapter batch sizes $\{1,2,4,8\}$.
AP can achieve up to $4.7\times$ speedup over FSDP\footnote{For per-adapter batch size $<4$, FSDP baseline cannot be used, so we estimate the throughput assuming it uses dummy data padding to batch size $=$ 4 (dashed bars).} and consistently outperforms all baselines.
AP's advantage is largest at small batch sizes (BS\,=\,1 to 2) where FSDP's all-reduce dominates; at BS\,=\,4 to 8, mLoRA and TP fall below FSDP while AP retains its edge.

\begin{figure}[t]
    \centering
    \includegraphics[width=\linewidth]{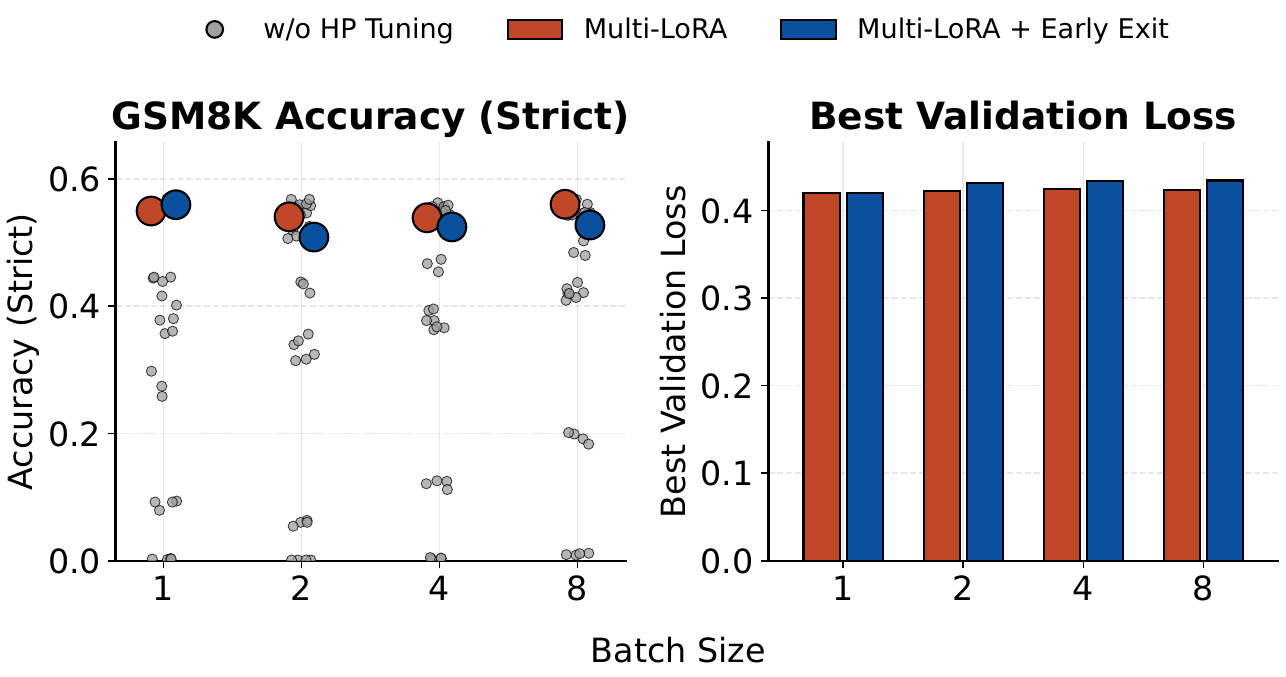}
    \caption{Single-GPU ablation on Llama-3.1-8B / GSM8K across per-adapter batch sizes. \textbf{Left:} GSM8K evaluation accuracy (strict parsing). Gray dots show individual adapter accuracies from the full sweep without selection. Colored markers compare the best accuracy found by Multi-LoRA batching alone vs.\ Multi-LoRA with early exit. \textbf{Right:} Best validation loss. Early exit preserves or improves quality by concentrating resources on promising configurations.}
    \label{fig:single-gpu-ablation}
\end{figure}

\begin{table}[t]
  \centering
  \small
  \caption{Kernel microbenchmark on Llama-3.2-1B, GSM8K, 32 adapters with ranks 16, 32, 64 equally mixed.
  \emph{PyTorch}: per-adapter LoRA run sequentially, backbone model passed in a batch.
  \emph{Sequential}: each adapter trained alone.
  \emph{Fused}: {\NAME}'s grouped Triton kernel.}
  \label{tab:kernel-micro}
  \scalebox{0.94}{
  \begin{tabular}{@{}crrrrrr@{}}
    \toprule
    Per-adapter & PyTorch & Sequential & Fused  & \multicolumn{2}{c}{Speedup over} \\
    \cmidrule(l){5-6}
    BS & (s) & (s) &  (s) & PyTorch & Sequential \\
    \midrule
    1 & 754.1  & 2016.6 & 394.6 & $1.91\times$ & $5.1\times$ \\
    2 & 514.4  & 1098.4 & 296.0 & $1.74\times$ & $3.7\times$ \\
    4 & 328.5  &  599.2 & 240.8 & $1.36\times$ & $2.5\times$ \\
    \bottomrule
  \end{tabular}}
\end{table}
\vspace{1pt}
\noindent \textbf{Quality impact of early exit.}
Figure~\ref{fig:single-gpu-ablation} isolates the quality impact of batched execution and early exit on Llama-3.1-8B with GSM8K.
Individual adapter accuracies exhibit high variance (gray dots), with many near-zero.
Batching finds strong configurations by training more candidates concurrently; adding early exit further improves the best result by concentrating resources on promising configurations.
Validation loss confirms no quality degradation.

\begin{figure}[t]
    \centering
    \includegraphics[width=0.99\linewidth]{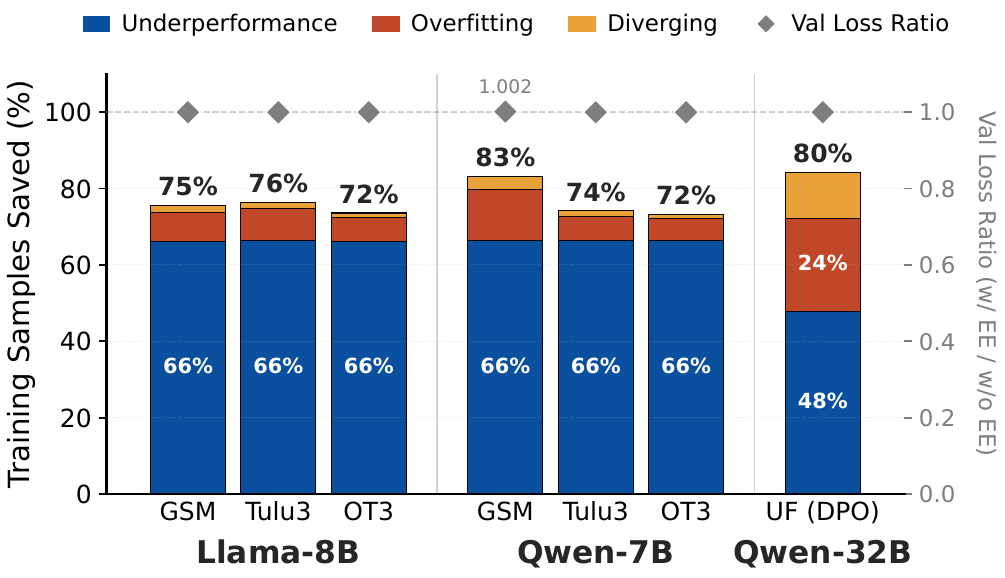}
    \vspace{5pt}
    \caption{Training samples saved by each early-exit pattern across seven model--dataset combinations. Stacked bars decompose savings by detector: underperformance filtering dominates in SFT (${\sim}66\%$), while overfitting and divergence contribute more in DPO.
    Green diamonds show the ratio of best validation loss w/ \textit{vs.}\ w/o early exit; values at or near 1.0 confirm that early termination preserves model quality.}
    \label{fig:samples-saved}
\end{figure}

\noindent \textbf{Early exit effectiveness.}
Figure~\ref{fig:samples-saved} reports the training samples saved by each early-exit pattern across Llama-3.1-8B and Qwen2.5-7B on three SFT datasets (GSM8K, Tulu-3, OpenThoughts3; 165--330 configurations each), plus Qwen2.5-32B on UltraFeedback with DPO (40 configurations). All use identical detector parameters ($w{=}2$, $p{=}2$, $\tau_{\text{gap}}{=}0.1$, $\tau_{\text{slope}}{=}0.001$, 5\% warmup, 25\% selection ratio). The three detectors collectively save 72--83\% of training samples. 
Underperformance filtering dominates in SFT (${\sim}66\%$ of saved samples), while overfitting and divergence contribute proportionally more in DPO (24\% and 12\%, respectively), reflecting DPO's more volatile loss dynamics. Notably, the same algorithm and thresholds generalize from SFT to DPO without modification, since both objectives produce training and evaluation losses on a comparable scale.

\section{Related Work}
\label{sec:related_work}
In this section, we give a comprehensive overview of prior efforts in three major aspects that closely related to our current work, including Hyperparameter Optimization, Multi-LoRA Serving and Training Systems.

\noindent \textbf{Hyperparameter Optimization.}
Hyperband~\cite{li2018hyperband} accelerates search through bandit-based resource allocation, progressively eliminating underperformers. ASHA~\cite{li2020asha} extends this with asynchronous successive halving for distributed settings. Early stopping methods complement resource allocation: Domhan et al.~\cite{domhan2015} extrapolate partial learning curves to terminate weak configurations early, while patience-based stopping~\cite{prechelt2002early} halts training when validation loss stagnates. {These techniques have not been studied in the context of LoRA fine-tuning, where short training durations and highly correlated adapter losses enable simpler yet effective early-exit strategies based on lightweight pattern detection.}

\vspace{2pt}
\noindent \textbf{Multi-LoRA Serving Systems.}
Multi-LoRA serving systems batch inference requests across adapters sharing a base model. S-LoRA~\cite{sheng2024slora} introduces unified paged memory management for thousands of concurrent adapters, Punica~\cite{chen2024punica} proposes the BGMV kernel for batched LoRA decoding, and dLoRA~\cite{wu2024dlora} dynamically migrates adapters across replicas for load balancing. {These systems target the memory-bandwidth-bound decoding regime; their kernels and parallelism strategies are not directly applicable to training, where the backward pass introduces gradient flow constraints and fundamentally different workload characteristics.}

\vspace{2pt}
\noindent \textbf{Multi-LoRA Training Systems.}
Recent works explore co-scheduling multiple LoRA adapters during training process. mLoRA~\cite{ye2024mlora} uses pipeline parallelism and a batched LoRA operator with graph-pruning optimizations to improve utilization. LoRAFusion~\cite{lorafusion} fuses memory-bound LoRA operations with compute-bound base GEMMs via Triton~\cite{tillet2019triton} kernels and introduces an adaptive MILP-based scheduler for balanced microbatches. PLoRA~\cite{yan2025plora} coordinates concurrent LoRA hyperparameter tuning jobs under shared hardware constraints. tLoRA~\cite{tlora} fuses adapters into an elastic super-model with rank-aware nano-batches and an online residual-capacity-aware scheduler to maximize collective throughput. {mLoRA~\cite{ye2024mlora} and LoRAFusion~\cite{lorafusion} both rely on pipeline parallelism, which suffers from workload imbalance across stages, even with careful scheduling. More fundamentally, all prior systems treat LoRA jobs as independent workloads; {\NAME} instead exploits the inter-job relationships inherent in hyperparameter tuning---using early-stage loss correlations to prune unpromising configurations and redirect resources to the most promising adapters.}
\section{Conclusion}
In this paper, we presented {\NAME}, a system that recasts multi-LoRA hyperparameter tuning as a unified system workload.
Three co-designed techniques drive its efficiency:
a loss-aware early-exit mechanism that detects diverging, overfitting, and underperforming configurations from loss trajectories;
a batched multi-LoRA engine with fused grouped GEMM kernels that amortizes base-model cost across co-located adapters;
and Adapter Parallelism, which assigns each GPU rank a distinct adapter set to eliminate idle ranks and redundant communication.
A hierarchical scheduler ties these components together, packing adapters within tasks and bin-packing tasks across a shared GPU cluster.
Our extensive experimental evaluation covers diverse LLMs, including popular Llama/Qwen models, various SFT datasets, and DPO algorithms, {\NAME} achieves up to $13.8\times$ speedup over state-of-the-art baselines while matching or exceeding expert-tuned adapter quality.

\pagestyle{plain}

\bibliographystyle{unsrt}
\bibliography{refs}



\clearpage
\appendix
\begin{figure*}
    \centering
    \includegraphics[width=\linewidth]{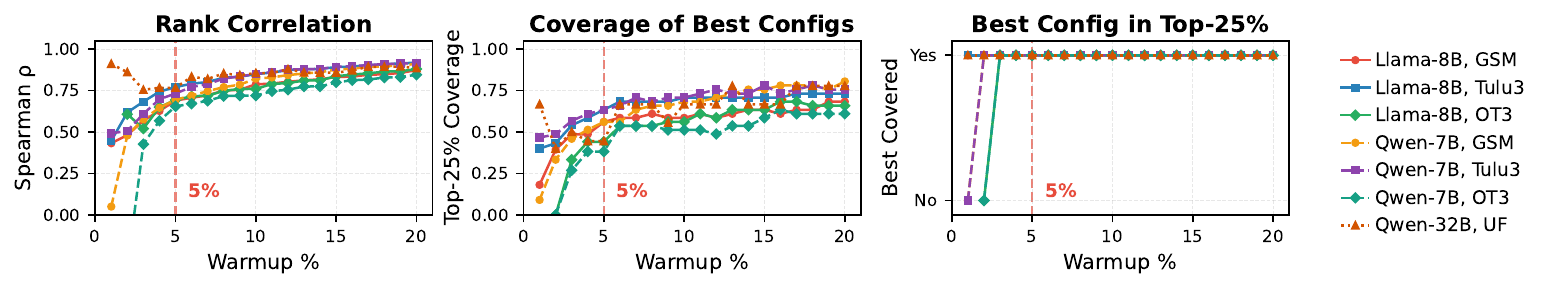}
    \caption{Sensitivity of early exit predictions to warmup percentage. \textbf{Left:} Spearman rank correlation between loss at warmup and final validation performance. \textbf{Middle:} Fraction of true top-25\% configurations captured by early predictions. \textbf{Right:}
  Whether the best configuration is included in the predicted top-25\%. The dashed line marks 5\%, our default warmup threshold, where all
  metrics stabilize across models and datasets.}
  \label{fig:sensitivity-sweep}
\end{figure*}

\section{Appendix}

\subsection{Grouped GEMM Kernel Design}
\label{sec:appendix-kernel}

Our Triton-based grouped GEMM kernel processes multiple LoRA adapters with heterogeneous batch sizes in a single kernel launch, without requiring data packing or token-level padding.

\textbf{Schedule-driven dispatch.}
Before each forward pass, the system builds a schedule table of \texttt{(adapter\_idx, block\_idx)} pairs on the CPU and transfers it asynchronously to the GPU.
Each Triton program reads its assignment from this table and computes the global token range from per-adapter \texttt{starts[i]} and \texttt{ends[i]} offset arrays.
Adapters with fewer tokens simply receive fewer thread blocks; boundary blocks apply a token-level mask (\texttt{offs\_m < end\_token}) so that no actual padding is needed in the activation tensor.
This design allows adapters with arbitrary per-adapter batch sizes to share a single flattened activation buffer and a single kernel launch, completely avoiding the need for sequence packing or batch-size alignment.

\textbf{Rank-only padding.}
When co-resident adapters use different LoRA ranks, the system pads only the weight matrices $\mathbf{A} \in \mathbb{R}^{d_\text{in} \times r_i}$ and $\mathbf{B} \in \mathbb{R}^{r_i \times d_\text{out}}$ along the rank dimension to $r_{\max} = \max_i r_i$, enabling them to be stacked into contiguous tensors of shape $[Z, d_\text{in}, r_{\max}]$ and $[Z, r_{\max}, d_\text{out}]$.
The padded columns are masked out inside the kernel (\texttt{offs\_r < R}), contributing zero to both GEMMs.
Because the rank dimension is typically small ($r \leq 128$), the padding overhead is negligible compared to approaches that pad along the token or batch dimensions.

\textbf{Fused base-output addition.}
The output of the LoRA path must be added to the base linear layer's output.
Our fused kernel variant merges this addition into the store phase of the second GEMM ($\mathbf{Y} = \mathbf{S}\mathbf{B} + \mathbf{Y}_\text{base}$), loading the base output tile directly inside the output loop and adding it in registers before the final store.
This eliminates one full read--write pass over the output tensor, yielding 3--36\% speedup depending on batch size.

\textbf{Grouped backward pass.}
Computing per-adapter weight gradients naively requires $2N$ independent GEMM launches (one $\nabla\mathbf{A}$ and one $\nabla\mathbf{B}$ per adapter).
We instead exploit the contiguous token layout: when all adapters share the same token count per adapter, we reshape the flattened activations into a batched 3-D tensor and use a single \texttt{torch.bmm} call for each of $\nabla\mathbf{A}$ and $\nabla\mathbf{B}$.
For the variable-token-count case, we use \texttt{torch.\_grouped\_mm} with cumulative offset descriptors.
Both paths reduce the backward to exactly two grouped kernel launches regardless of the number of adapters.

\textbf{Homogeneous batch grouping.}
When adapter parallelism shards the LoRA computation across multiple GPUs, each rank must process the same token count per adapter to avoid communication mismatches during gradient allreduce.
The intra-task scheduler therefore groups adapters by per-adapter batch size and co-locates same-batch-size jobs on the same executor (\S\ref{sec:appendix-intra-task}).
This policy simultaneously ensures correctness under adapter parallelism and maximizes the fraction of forward passes that hit the more efficient \texttt{bmm}-based backward path.


\subsection{Warmup Hyperparameter Sensitivity Evaluation}
\label{sec:appendix-sensitivity}
\textbf{Warmup threshold.}
  Figure~\ref{fig:sensitivity-sweep} analyzes how the warmup period affects the reliability of early exit decisions. We vary the warmup
  percentage from 1\% to 20\% and measure three metrics: (1) Spearman rank correlation between early-stage loss and final validation
  performance, which indicates the correlation between the early performance and the final performance, (2) coverage of the true top-25\% configurations when selecting based on early observations, and (3) whether the globally
  best configuration falls within the predicted top-25\%. Across seven model-dataset combinations, rank correlation stabilizes above 0.7 by
   5\% warmup, and top-25\% coverage reaches 60--80\%. Crucially, the best configuration is reliably captured within the top quartile at
  5\% for all settings. These results justify our default 5\% warmup threshold: shorter warmups yield unreliable rankings (e.g., Spearman
  $\rho < 0.5$ at 2\%), while longer warmups delay early exit decisions without improving prediction quality.

\subsection{Intra-Task Scheduling Details}
\label{sec:appendix-intra-task}

\textbf{Memory profiling.}
The memory profiler operates in two phases. First, it fixes the number of adapters to $N{=}1$ and performs a binary search over the total batch size $B$ to find the largest $B_{\max}$ that fits within a safety margin of the GPU's HBM capacity. Second, it sweeps over a grid of configurations $(N, b)$ where $b \in \{1, 2, 4, 8, 16, 32\}$ and $N \cdot b \leq B_{\max}$, measuring empirical peak memory for each configuration by creating synthetic data, running a single training step, and recording \texttt{torch.cuda.max\_memory\_reserved}. The collected data points are fitted to a linear regression model $\hat{M}(B) = k_0 + k_1 B L$ (where $L$ is the sequence length), which the scheduler queries at runtime. When adapter parallelism is active, profiling runs per-rank.

\textbf{Admission and backfill policy.}
The scheduler groups training jobs by their per-adapter batch size and co-locates same-batch-size jobs on the same executor, aligning with the grouped GEMM kernel. For each proposed adapter admission, the scheduler queries $\hat{M}(B_{\text{current}} + b_{\text{new}})$ and admits the adapter only if the predicted peak memory falls within the safety margin. Adapters are admitted greedily in decreasing batch-size order until no further admission is feasible. When a training job exits via early exit or normal completion, the scheduler first attempts to fill the vacated slot with another pending job of the \emph{same} batch size, preserving the executor's homogeneous packing. If no same-batch-size jobs remain in the queue, the scheduler may admit a job with a different batch size, provided the memory model confirms the mixed configuration fits. This policy keeps executors saturated while maximizing grouped GEMM efficiency: batch-size homogeneity is preferred but not enforced, and the system gracefully degrades to mixed packing as the candidate pool shrinks.

\subsection{Training Details}
\label{sec:appendix-training-details}

\textbf{Hyperparameter search space.}
For single-GPU models (7B--8B scale), we search over 60 configurations:
learning rates $\in \{$1e-5, 5e-5, 2e-4, 3e-4, 5e-4$\}$,
LoRA ranks $\in \{16, 32, 64\}$,
and per-adapter batch sizes $\in \{1, 2, 4, 8\}$.
For multi-GPU models (32B--70B scale), we search over 64 configurations:
learning rates $\in \{$1e-5, 5e-5, 1e-4, 3e-4$\}$,
LoRA ranks $\in \{16, 32, 64, 128\}$,
and batch sizes $\in \{1, 2, 4, 8\}$.
All experiments train for 3 epochs with gradient accumulation set to 1, gradient checkpointing enabled, constant learning rate schedule, and paged AdamW 8-bit optimizer with weight decay 0.01.
LoRA adapters are applied to all attention and MLP projection layers (q, k, v, o, gate, up, down) with $\alpha = 2r$.

\end{document}